%% file: template.tex
\RequirePackage{fix-cm}
\documentclass[twocolumn]{svjour3}          
\smartqed  
\usepackage{graphicx}
\usepackage{enumitem}
\usepackage{caption}
\usepackage{subcaption}
\usepackage[table]{xcolor}
\usepackage{array}
\usepackage{multirow}
\usepackage{booktabs}
\usepackage{threeparttable}
\usepackage{amsmath}

\usepackage{pifont}

\newcommand{\none}{—}

\usepackage{fontawesome5}
\usepackage[pagebackref=true,breaklinks=true,colorlinks,bookmarks=false]{hyperref}

\definecolor{mygray}{gray}{.90}
\definecolor{mylgray}{gray}{.70}
\definecolor{reda}{RGB}{202,0,0}
\definecolor{redb}{RGB}{217,148,143}
\definecolor{lightred}{RGB}{125,0,0}
\definecolor{myyellow}{RGB}{190,144,0}
\definecolor{mygreen}{RGB}{0,136,51}
\definecolor{mylgreen}{RGB}{0,128,0}
\definecolor{myblue}{RGB}{0,102,204}

\newcommand{\basevalue}[1]{{\textcolor{mylgreen}{#1}}}

\newcommand{\parhead}[1]{\noindent\textbf{#1} }
\newcommand{\PromptEverything}{\faBuromobelexperte}
\newcommand{\PromptPoint}{\faDotCircle[regular]}
\newcommand{\PromptMask}{\faDisease}
\newcommand{\PromptBox}{\faBorderStyle}
\newcommand{\PromptICL}{\faDice}

\begin{document}

\title{Inspiring the Next Generation of Segment Anything Models: 
	Comprehensively Evaluate SAM and SAM 2 with Diverse Prompts Towards Context-Dependent Concepts under Different Scenes}

\titlerunning{Comprehensive Evaluation of SAMs on CD Concepts}        

\author{Xiaoqi Zhao$^{1}$ \and
        Youwei Pang$^{2}$  \and
        Shijie Chang$^{3}$ \and
        Yuan Zhao$^{3}$ \and \\
        Lihe Zhang$^{3}$ \and
        Chenyang Yu$^{3}$ \and
        Hanqi Liu$^{4}$ \and
        Jiaming Zuo$^{4}$ \and \\
        Jinsong Ouyang$^{1}$ \and
        Weisi Lin$^{2}$ \and
        Georges El Fakhri$^{1}$ \and\\
         Huchuan Lu$^{3}$ \and
        Xiaofeng Liu$^{1}$ \\
         {\tt \url{https://github.com/lartpang/SAMs-CDConcepts-Eval}}
}


\institute{
$^{1}$  Yale University, USA\\
$^{2}$  Nanyang Technological University, Singapore\\
$^{3}$  Dalian University of Technology, China\\
$^{4}$ X3000 Inspection Co., Ltd, China\\
\\
Lihe~Zhang is the corresponding author.
\\
Youwei Pang is the project leader. 
\\
Shijie Chang and Yuan Zhao are the core maintainers of the project repository/code.
\\
\\
E-mail: \\
xiaoqi.zhao@yale.edu \\
zhaoxq.cv@gmail.com \\
zhanglihe@dlut.edu.cn \\
1artpang@gmail.com \\
xiaofeng.liu@yale.edu\\
\\
This work was supported by the National Natural Science Foundation of China under Grant 62431004 and 62276046.
}

\date{Received: date / Accepted: date}

\maketitle
\begin{abstract}
As large-scale foundation models trained on billions of image–mask pairs covering a vast diversity of scenes, objects, and contexts, SAM and its upgraded version, SAM 2, have significantly influenced multiple fields within computer vision. Leveraging such unprecedented data diversity, they exhibit strong open-world segmentation capabilities, with SAM 2 further enhancing these capabilities to support high-quality video segmentation. 
While SAMs (SAM and SAM 2) have demonstrated excellent performance in segmenting context-independent concepts like people, cars, and roads, they overlook more challenging context-dependent (CD) concepts, such as visual saliency, camouflage, industrial defects, and medical lesions. CD concepts rely heavily on global and local contextual information, making them susceptible to shifts in different contexts, which requires strong discriminative capabilities from the model. 
The lack of comprehensive evaluation of SAMs limits understanding of their performance boundaries, which may hinder the design of future models. In this paper, we conduct a thorough evaluation of SAMs on 11 CD concepts across 2D and 3D images and videos in various visual modalities within natural, medical, and industrial scenes. We develop a unified evaluation framework for SAM and SAM 2 that supports manual, automatic, and intermediate self-prompting, aided by our specific prompt generation and interaction strategies. We further explore the potential of SAM 2 for in-context learning and introduce prompt robustness testing to simulate real-world imperfect prompts.
Finally, we analyze the benefits and limitations of SAMs in understanding CD concepts and discuss their future development in segmentation tasks. This work aims to provide valuable insights to guide future research in both context-independent and context-dependent concepts segmentation, potentially informing the development of the next version — SAM 3.
\end{abstract}

\begin{figure*}[!t]
  \centering
  \includegraphics[width=\linewidth]{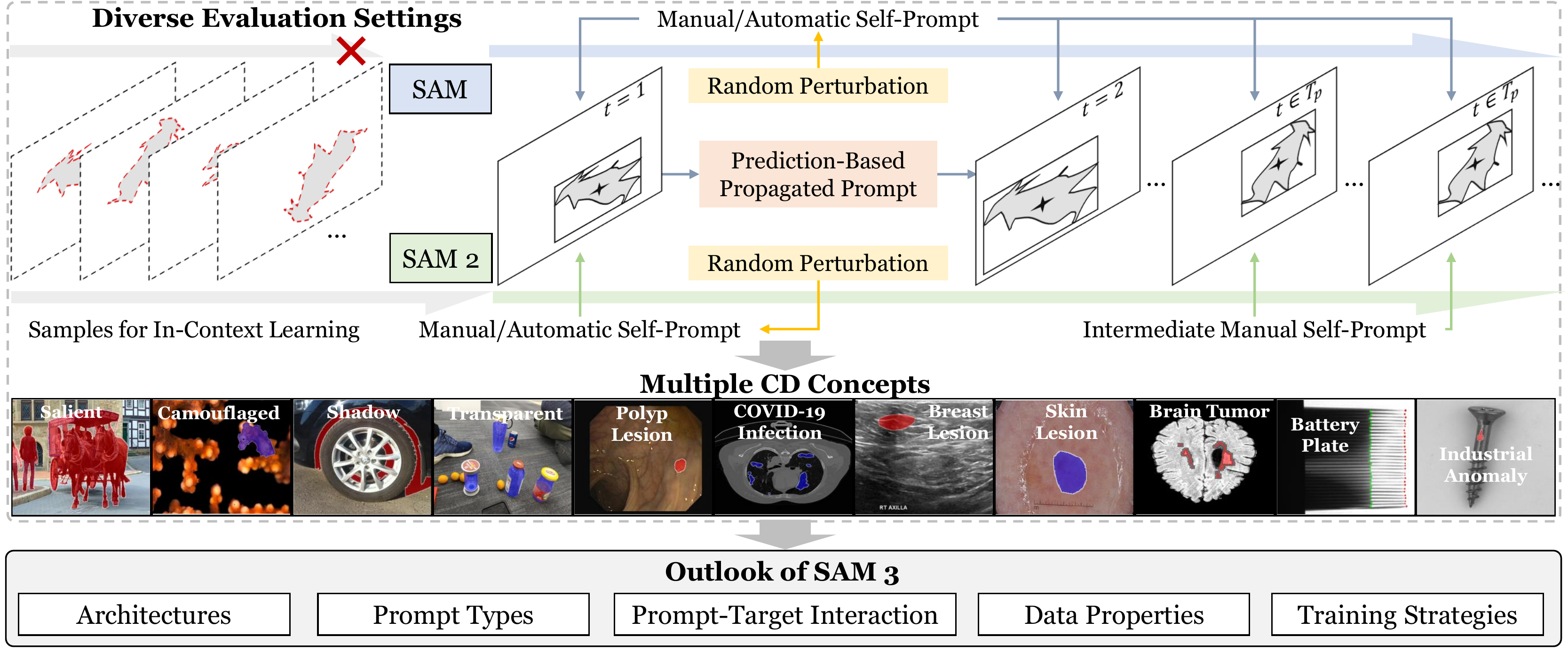}
  \caption{Organization. (1) A unified evaluation framework for SAM and SAM 2; (2) Comprehensive evaluation for 11 different context-dependent concepts; (3) Outlook of SAM 3.}
  \label{fig:teaser}
\end{figure*}

\begin{table*}[!ht]
  \centering
  \caption{Summary of the characteristics in different evaluation works. Different prompt types:
    \PromptEverything: Everything;
    \PromptMask: Mask;
    \PromptBox: Box;
    \PromptPoint: Point;
    \PromptICL: In-Context Learning. 
    ``+PR'': Prompt Robustness Analysis.
  }
  \label{tab:survey_comparison}
  \resizebox{\linewidth}{!}{\input{table/survey_comparison.tex}}
\end{table*}

\section{Introduction}
As a foundation model in the field of image segmentation, Segment Anything Model (SAM~\cite{SAM}) has demonstrated remarkable performance across various scenarios, spurring research interest in unified/generalist models~\cite{CDCU-Spider,Painter,MedSAM}, in-context visual learning~\cite{SegGPT,UniverSeg,Matcher}, and SAM-adaptors~\cite{HQSAM,MedSAMAdapter,SAMAdaptor2}. Recently, the upgraded version, SAM 2~\cite{SAM2}, has introduced powerful video object segmentation capabilities, expected to ignite a new wave of research.

In philosophy and cognitive science~\cite{CICD_1}, the concept of an object is typically divided into context-independent (CI) and context-dependent (CD) concepts. Recently, Zhao \textit{et al.}~\cite{CDCU-Spider} first provide a detailed distinction of CI and CD concepts within the image segmentation field. Traditional semantic segmentation datasets~\cite{COCO-Stuff,Cityscapes} usually focus on the CI concepts such as roads, vehicles, and people that are relatively easy to segment. Regardless of the environment, the shape and category of these objects are stable, allowing models to focus solely on the intrinsic features of the objects for effective segmentation.
In real-world scenarios, predictions of CI concepts often serve as preliminary steps for further scene analysis. Different from them, CD concept segmentation tasks are explicitly oriented towards functional applications, demonstrating direct value in visual attention perception, medical lesion segmentation, and industrial inspection. However, due to the environmental dependence, concept variability, and scene specificity, existing CD concepts methods often rely on domain-specific specialized models, making unified CD concept segmentation more challenging. Can SAMs perfectly segment CD concepts? Existing works have evaluated the segmentation performance of SAMs on saliency~\cite{Lian_SAM_report}, camouflage~\cite{Ji_SAM_report1}, shadow~\cite{Chen_SAM_report}, and colon polyps~\cite{Zhou_SAM_report}. As shown in Tab.~\ref{tab:survey_comparison}, these evaluations are too domain-specific rather than the high-level CD concepts perspective. Most of these studies are limited to quantitative evaluations on a small set of datasets under the everything prompt mode. Compared with them, we have obvious advantages in the evaluation breadth and depth of scenarios, CD concepts, datasets, modalities, and prompting types. We believe that to fairly assess the capability of SAMs in CD concepts segmentation, it is essential to conduct enough experiments on diverse concepts and benchmarks, as well as a variety of prompt types and strategies. Insufficient experimentation can easily introduce bias and lead to subjective conclusions.

The organization of this paper is illustrated in Fig.~\ref{fig:teaser}. \textbf{First}, we design a unified evaluation framework for SAMs, integrating manual, automatic, and intermediate manual self-prompting methods. Everything, point, and box prompts naturally fall within this comprehensive scope. Notably, we develop a prediction-based propagated prompt and non-current sample prompting for in-context learning inference mode, targeting the serialization predictions and memory attention characteristics of SAM 2. \textbf{Next}, we conduct quantitative experiments on image segmentation in both basic and in-context learning modes, as well as video and 3D segmentation across 33 datasets covering 11 CD concepts. \textbf{Finally}, we conduct an in-depth analysis of current representative unified segmentation models in terms of architecture, prompt types, prompt-target interactions, training data, and strategies to inspire the next generation of Segment Anything Models.

\begin{figure*}[!t]
  \centering
  \includegraphics[width=0.96\linewidth]{figure/CI_vs_CD_concepts.pdf}
  \caption{Visual comparison between context-independent concepts and context-dependent concepts.}
  \label{fig:CI_vs_CD_concepts}
\end{figure*}

Our main contributions can be summarized as follows: 

{\textbf{\uppercase\expandafter{\romannumeral1}) {Motivation and significance.}}} While SAMs have been primarily evaluated on context-independent (CI) tasks, their behavior on CD concepts such as saliency, camouflage, medical abnormalities, and industrial defects remains underexplored. To our knowledge, this study  provides the first large-scale and unified evaluation of SAMs on CD segmentation tasks, helping the community better understand their performance boundaries and limitations. This can guide future model design and deployment strategies.

{\textbf{\uppercase\expandafter{\romannumeral2}) {Comprehensive evaluation modes.}}} We propose six evaluation settings for SAMs, including Everything, Mask, Box, Point, In-Context Learning, and Prompt 
Robustness. The last two are introduced for the first time in the context of SAM evaluations and reflect important aspects of real-world usage.

{\textbf{\uppercase\expandafter{\romannumeral3})  {Broad coverage of domains and data types.}}} Our evaluation spans 33 datasets across 11 CD concepts and includes natural, medical, and industrial scenarios. These are evaluated on 2D images, videos, and 3D medical volumes. To the best of our knowledge, this represents the most comprehensive CD-oriented evaluation of SAMs to date.

{\textbf{\uppercase\expandafter{\romannumeral4})  {Unified prompting strategies for heterogeneous data.}}} We design three prompting strategies to enable cross-domain evaluation, including prediction-based prompt propagation, bidirectional inference, and in-context learning for multi-frame and volumetric data. These tools ensure that SAMs can be applied in open-world scenarios across diverse settings without task-specific retraining.

{\textbf{\uppercase\expandafter{\romannumeral5})  {Design insights for future models.}}} By analyzing the strengths and weaknesses of SAMs on CD concepts and comparing them with other unified models, we can provide meaningful outlooks for the next generation of SAM.

{\textbf{\uppercase\expandafter{\romannumeral6})  {Practical benchmarks and tools.}}} We provide complete baseline results and unified interfaces that can serve as a standardized reference for future research. This makes it easier for researchers to evaluate new models on a variety of CD tasks and contributes to reproducibility and accessibility.

\section{Related Works}
\subsection{Context-Independent Concepts vs. Context-Dependent Concepts}

The distinction between context-independent (CI) and context-dependent (CD) concepts originates from philosophy of language and cognitive science, where it has long been studied as a fundamental issue in conceptual representation~\cite{CICD_1,CICD_2,CICD_3}. The key difference lies in whether contextual information plays a decisive role in defining a concept. While CI concepts are relatively stable and environment-invariant, CD concepts are inherently determined by their surrounding spatial or semantic context.
This distinction has recently been introduced into the computer vision community. In particular, Spider~\cite{CDCU-Spider} explicitly incorporates CD concepts into segmentation benchmarks and proposes a unified modeling framework for them. 
Fig.~\ref{fig:CI_vs_CD_concepts} illustrates the comparison between  CI and CD concepts in vision. 
For CI concepts, such as ``cars,'' ``bears,'' or ``circles,'' the visual identity remains stable across different environments, enabling models to segment them primarily based on intrinsic appearance and structure cues. By contrast, CD concepts lose their meaning when isolated from their context. For instance, saliency depends on visual attention mechanisms that consider relationships among objects, shadows are defined by the physical projection of their casting objects, medical lesions become identifiable only under specific imaging conditions relative to surrounding healthy tissue, and industrial defects can be determined only in comparison with the main body of the product. 
Therefore, compared to CI concepts, the recognition and understanding of CD concepts are significantly more challenging, mainly due to their three unique characteristics: \textit{{\uppercase\expandafter{\romannumeral1}) {environmental dependence}}}, requiring contextual cues for recognition; \textit{{\uppercase\expandafter{\romannumeral2}) \textit{concept variability}}}, as their definitions may change with background or modality (e.g., a salient object under one condition may become camouflaged under another); \textit{{\uppercase\expandafter{\romannumeral3}) \textit{scene specificity}}}, as they are often meaningful only within particular domains such as medical imaging or industrial inspection. These properties make CD concepts segmentation a more complex and underexplored problem, calling for models that can capture contextual dependencies beyond object appearance alone.

\subsection{Context-Dependent Concepts Segmentation}
Context-dependent (CD) concept segmentation has garnered significant attention over the years. These concepts rely on specific spatial contexts to define the concepts of interest, posing unique challenges and driving advanced designs for specialized models.
\noindent\textit{\textbf{\uppercase\expandafter{\romannumeral1}) Background Complexity and Similarity.}} In tasks like camouflaged and transparent object segmentation, highly similar backgrounds make it difficult for the model to distinguish between the target object and surroundings. This requires models with enhanced background understanding and segmentation capabilities~\cite{SINet-v2_COD,Zoomnext,rumaksari2017background}. \noindent\textit{\textbf{\uppercase\expandafter{\romannumeral2}) Object Boundary Ambiguity.}} In tasks such as transparent object and medical lesion segmentation, smooth transitions between the object and surroundings often lead to boundary ambiguity. Models can missegment these fuzzy edges, necessitating precise boundary recognition and shape modeling capabilities~\cite{PraNet,MSNet,TOM-Net_Transparent}. \noindent\textit{\textbf{\uppercase\expandafter{\romannumeral3}) Context Dependency.}} Models need strong context-awareness, adjusting segmentation strategies based on the surrounding environment rather than relying solely on local features of the target objects~\cite{RCARP_Glass,BCS-Net_COVID-19,GCPANet}.

\subsection{Unified Multi-Concept Segmentors}
The development of large foundation models and visual prompt technology has led to the emergence of various models aimed at achieving AGI, notably in unified and generalist segmentation. Over the past year, SAM has become a standout segmentor due to its simple architecture, extensive data training, and impressive performance. Following SAM, more generalist models aim to accurately segment context-independent concepts with different prompt learning strategies. UniverSeg~\cite{UniverSeg} excels in unifying medical image segmentation across diverse tasks with domain-agnostic representations. SegGPT~\cite{SegGPT} employs flexible, prompt-based segmentation using transformer architecture, while HQSAM~\cite{HQSAM} produces high-quality, high-resolution masks with strong generalization and real-time inference. For context-dependent concepts, EVP~\cite{EVP} enhances low-level structure segmentation through explicit visual prompting, while GateNet~\cite{GateNetv2} offers a versatile gated network for various CD concepts tasks. Spider~\cite{CDCU-Spider} and VSCode~\cite{VSCode} leverage 2D prompt learning to understand background-foreground relationships. Recently, SAM 2 built on SAM by introducing memory attention and multiple frame prompts, utilizing large video datasets to advance video object segmentation. Its approach is expected to invigorate 3D, video, and few-shot/co-segmentation fields.

\subsection{SAMs Evaluation}
The development of any technology inherently presents a dual nature.
On one hand, SAMs, as segmentation foundation models, provide significant potential for direct application across tasks. On the other hand, SAMs challenge the long-standing independence of specialized segmentation sub-fields, raising the question, ``Is segmentation as we know it obsolete?'' Existing reports have focused on tasks like camouflaged object detection (COD)~\cite{Ji_SAM_report1}, shadow detection~\cite{Ji_SAM_report2}, polyp segmentation~\cite{Zhou_SAM_report}, and underwater salient object detection~\cite{Lian_SAM_report}. Following the trend in unified/specialist segmentation methods, which categorizes segmentation into context-independent (CI) and context-dependent (CD) concepts, we aim to provide a fair and comprehensive evaluation of SAMs' performance across various CD concepts. The goal is to establish an evaluation baseline for future research, minimizing redundant work.

\section{Experiments}
\subsection{Datasets}
\label{app:sec:datasets}

\begin{table*}[!t]
  \centering
  \caption{Dataset links.}
  \label{tab:dataset_link}
  \resizebox{\linewidth}{!}{\input{table/dataset_link}}
\end{table*}

\subsubsection{Natural Scene Data}

\noindent\textbf{Image Salient Object Detection.}
DUTS~\cite{DUTS} consists of 10,553 training images and 5,019 testing images, covering diverse scenes with high-quality pixel-level saliency annotations and widely used for evaluating salient object detection models. 
ECSSD~\cite{ECSSD} includes 1,000 images containing complex scenes where salient objects often blend into the background, challenging models to differentiate salient regions. 
DUT-OMRON~\cite{DUT-OMRON} features 5,168 images with complex backgrounds and small objects, making it an essential dataset for assessing robustness of salient object detection algorithms. 
HKU-IS~\cite{HKU-IS} contains 4,447 images (2500 for training, 500 for validation, and 1447 for testing) with detailed edge annotations, focusing on large salient objects with clear boundaries, which challenge models to capture fine-grained details. 
PASCAL-S~\cite{PASCAL-S} is derived from PASCAL VOC 2010 with 850 images annotated by multiple experts, aiming to test saliency models in natural and complex scenes. 
In our experiments, the testing data is from ECSSD, DUT-OMRON, PASCAL-S, and the testing sets of DUTS and HKU-IS.

\noindent\textbf{Video Salient Object Detection.}
DAVIS16~\cite{DAVIS16} is a benchmark dataset comprising 50 high-quality video sequences (30 for training and 20 for validation) with pixel-level annotations for object segmentation in dynamic scenes.
The dataset is characterized by complex settings, including frequent occlusions, fast motion, and intricate backgrounds, making it a popular choice for evaluating video object segmentation models.
In our experiments, we use the validation subset of DAVIS16, which contains 20 sequences specifically selected to assess the generalization performance of models in diverse and challenging scenarios.
DAVSOD~\cite{DAVSOD} is a large-scale video saliency detection dataset that includes 226 video sequences (61 for training, 46 for validation, and 80 for testing) with pixel-level saliency annotations.
It is designed to evaluate models in a wide range of scenarios, including dynamic scenes and camouflaged objects, providing a comprehensive benchmark for saliency detection tasks.
The testing set is divided into three splits based on difficulty levels: DAVSOD$_E$ (easy, 35 sequences), DAVSOD$_N$ (normal, 25 sequences), and DAVSOD$_H$ (hard, 20 sequences).
These splits allow for a nuanced evaluation of model performance under varying complexities, ranging from relatively straightforward scenes to highly intricate and visually challenging scenarios.
In our study, we use all three splits to comprehensively analyze the model's robustness and adaptability across different difficulty levels.

\noindent\textbf{Image Camouflaged Object Detection.}
Following the recent typical methods~\cite{Zoomnext,COD-SARNet}, we adopt a similar evaluation strategy for the CAMO~\cite{CAMO}, COD10K~\cite{SINet_COD}, and NC4K~\cite{Rank-Net_COD} datasets.
Specifically, for CAMO, we use a subset containing 1,250 images that include camouflaged objects.
CAMO is a specialized dataset designed to evaluate the detection of camouflaged objects in complex backgrounds, featuring a diverse range of challenging scenarios where objects are intentionally concealed within their surroundings. 
For COD10K, we focus on a subset of 5,066 images that are carefully selected to include scenes with camouflaged objects.
This subset is annotated with pixel-level ground truth, providing a comprehensive benchmark for evaluating model performance in detecting objects that seamlessly blend into diverse and complex natural environments. 
For NC4K, we use the entire dataset comprising 4,121 high-resolution images.
NC4K is specifically curated to assess model generalization capabilities by presenting camouflaged objects across a wide variety of natural scenes with intricate details and challenging visual conditions. 
Notably, in our experiments, we follow the common practice of only testing on images containing camouflaged objects for CAMO and COD10K, while using the entire NC4K dataset as the test set.
This evaluation protocol ensures consistency with recent works and provides a fair comparison of model performance. 

\noindent\textbf{Video Camouflaged Object Detection.}
CAD~\cite{VCOD-CAD}, consisting of 9 sequences, focuses on camouflaged object detection in continuous video frames, emphasizing the temporal consistency of models in dynamic backgrounds.
MoCA-Mask~\cite{VCOD-MoCA-Mask} features high-quality mask annotations for camouflaged objects in videos, challenging models to detect and segment targets in motion.
It is divided into two subsets: 71 sequences for training and 16 sequences for testing.
The entire CAD dataset and the testing subset of MoCA-Mask are used for evaluating the methods.

\noindent\textbf{Image Shadow Detection.}
SBU~\cite{SBU} is a widely-used benchmark dataset for shadow detection.
Its testing set contains 700 outdoor scene images with pixel-level shadow annotations.
These images include diverse scenarios with shadows cast by various objects, providing a robust basis for evaluating the precision of shadow detection models.
ISTD~\cite{ISTD} contains 1,870 sets of images, each consisting of a shadow image, a shadow-free counterpart, and a shadow mask.
This dataset is specifically designed for both shadow detection and removal tasks.
It is randomly split into 1,330 image sets for training and 540 image sets for testing.
In our experiments, we directly use the testing sets of both SBU and ISTD, ensuring consistency with prior works.

\noindent\textbf{Video Shadow Detection.}
VISAD~\cite{VISAD-DS} is a comprehensive video shadow detection dataset comprising 81 videos, curated from various public video datasets to address the challenges of detecting shadows in dynamic scenarios.
The dataset is divided into two subsets based on scene types: the Driving Scenes (VISAD-DS) subset and the Moving Object Scenes (VISAD-MOS) subset, denoted as DS and MOS, respectively.
This division enables targeted evaluation of shadow detection models across distinct real-world settings.
VISAD-DS focuses on driving scenarios, featuring videos captured in urban streets, highways, and similar environments.
Shadows in this subset are typically caused by moving vehicles, pedestrians, and static objects such as buildings and trees.
The interplay of dynamic elements and structured backgrounds makes this subset a challenging benchmark, particularly for detecting shadows that may overlap with objects of interest or blend with road features.
VISAD-MOS highlights scenes dominated by moving objects, such as people, animals, or vehicles, in more diverse and unstructured environments.
Shadows in these videos are influenced by variable lighting conditions and intricate object interactions, making it a critical test bed for evaluating models' ability to generalize across complex scenarios.
By utilizing both VISAD-DS and VISAD-MOS in our experiments, we aim to comprehensively evaluate the performance of shadow detection models across a wide spectrum of dynamic and challenging settings.
This approach ensures a robust assessment of models' adaptability to diverse scene characteristics.

\noindent\textbf{Transparent Object Segmentation.}
Trans10K~\cite{TransLab_Transparent} contains 10,428 images designed for pixel-level segmentation of transparent objects, which are challenging due to their translucent nature.

\subsubsection{Medical Scene Data}

\noindent\textbf{Image Polyp Lesion Segmentation.}
There are five popular polyp segmentation datasets.  
Kvasir~\cite{Kvasir} contains 1,000 colonoscopy polyp images. 
ETIS~\cite{ETIS} with 196 high-resolution images focus on detecting small polyps.
CVC-ClinicDB~\cite{CVC-ClinicDB} with 612 colonoscopy images commonly used for segmentation evaluation.  
CVC-ColonDB~\cite{CVC-ColonDB} comprising 380 images with complex backgrounds challenging models to detect small objects, 
and Endoscene~\cite{Endoscene} with 912 images covering diverse polyp detection scenarios. 
Since these datasets are small in data-scale, in order to avoid performance volatility as much as possible, we follow the Spider~\cite{CDCU-Spider} to calculate the average results of the five datasets for performance evaluation.

\noindent\textbf{Video Polyp Lesion Segmentation.}
CVC-612-T, CVC-612-V, and CVC-300-TV~\cite{PNS-Net_Polyp} focus on polyp segmentation in video sequences, considering temporal continuity and dynamics.

\noindent\textbf{Skin Lesion Segmentation.}
ISIC-2018~\cite{ISIC18}, the ISIC challenge dataset, contains over 13,000 annotated skin lesion images, primarily for melanoma detection.

\noindent\textbf{COVID-19 Lung Infection Segmentation.}
COVID-19 CT~\cite{Inf-Net} comprises CT scans of patients with pixel-level lung infection annotations.

\noindent\textbf{Brain Tumor Segmentation.}
BraTS2020~\cite{BraTS2020} is an MRI-based dataset providing multi-modal annotations (T1, T2, T1ce, Flair) for brain tumor segmentation. 
ISBI2015~\cite{ISBI2015} targets multiple sclerosis lesion segmentation; we specifically use the Flair modality for lesion analysis.

\noindent\textbf{Breast Lesion Segmentation.}
BUSI~\cite{BUSI} is an ultrasound dataset with pixel-level annotations for breast lesions, widely used in breast cancer detection research.

\subsubsection{Industrial Scene Data}

\noindent\textbf{Power Battery Detection.}
PBD~\cite{AI4Industry-PBD} is an X-ray dataset used for detecting defects in power batteries, particularly focusing on internal structural issues.

\noindent\textbf{Surface Anomaly Detection.}
MVTec-AD~\cite{MVTec-AD} includes high-resolution images across 15 categories of industrial products used for detecting surface defects. 
VisA~\cite{VisA} covers various industrial products for visual anomaly detection, focusing on defect detection across different materials. 
BTAD~\cite{BTAD} contains ultra-high-resolution images for detecting small defects on metal surfaces.

\subsection{Evaluation Metrics}
\label{app:sec:metrics}
For salient object detection (SOD) and camouflaged object detection (COD), we utilize weighted F-measure (\(F^{\omega}_{\beta}\))~\cite{Fwb}, S-measure ({Sm})~\cite{S-m}, and mean absolute error (MAE).
The weighted F-measure accounts for spatial significance by assigning greater weight to pixels in critical regions, thereby providing a more precise balance between precision and recall.
The S-measure emphasizes structural similarity between the predicted saliency map and the ground truth, combining object-aware and region-aware evaluations to capture holistic accuracy.
Meanwhile, the MAE serves as a straightforward metric to calculate the pixel-wise average absolute difference, offering an intuitive measure of overall prediction accuracy without considering spatial structure.  

For shadow detection (SD) and transparent object segmentation (TOS), where significant class imbalance often exists, we adopt the balanced error rate (BER)~\cite{BER}. This metric is defined as the average of the false positive rate (FPR) and false negative rate (FNR), ensuring fair evaluation across imbalanced datasets by equally penalizing errors in positive and negative classes.

In medical lesion object segmentation (LOS), we report the mean Intersection over Union (mIoU) together with the foreground Dice similarity coefficient. 
mIoU provides a complementary perspective by averaging IoU across both lesion (foreground) and background regions, offering a more balanced view of segmentation performance. The Dice coefficient directly reflects the accuracy of lesion region prediction by emphasizing the overlap between predicted and ground-truth lesion masks, which is particularly suitable for evaluating small or irregular targets.

For power battery detection (PBD), both spatial accuracy and numerical correctness are critical. Location mean absolute error (AL-MAE, CL-MAE, OH-MAE)~\cite{AI4Industry-PBD} measures the mean absolute error of detected locations under different configurations, such as alignment, clustering, and outlier handling.
Additionally, point number accuracy (PN-ACC) evaluates the ability to predict the number of detected power battery units, ensuring reliability in industrial applications where both precision and count are essential.

In the domain of surface anomaly detection (AD), a combination of image-level and pixel-level metrics is utilized. 
At the image level, we employ I-AUROC and I-AP to assess classification performance, with the former measuring the area under the receiver operating characteristic curve and the latter summarizing precision-recall tradeoffs. 
At the pixel level, P-AUROC and P-AP provide analogous measures for segmentation performance, focusing on accurate localization of anomalous regions. 
Additionally, P-PRO quantifies per-region overlap between predicted and ground-truth anomalous regions, offering a fine-grained evaluation of segmentation quality that emphasizes spatial alignment.

These metrics collectively provide a comprehensive framework for evaluating model performance across diverse tasks, addressing challenges such as class imbalance, structural similarity, spatial alignment, and numerical accuracy.
By leveraging these tailored evaluation measures, we ensure a rigorous and fair assessment of the SAMs under various application scenarios.  
The implementation of all evaluation metrics is based on the publicly available toolkits. \footnote{
\url{https://github.com/Xiaoqi-Zhao-DLUT/PySegMetric\_EvalToolkit}\\
\url{https://github.com/zhaoyuan1209/PyADMetric\_EvalToolkit}\\
\url{https://github.com/Xiaoqi-Zhao-DLUT/X-ray-PBD}} 

\subsection{Implementation Details}
\label{sec:Implementation}
\begin{figure}
  \centering
  \includegraphics[width=\linewidth]{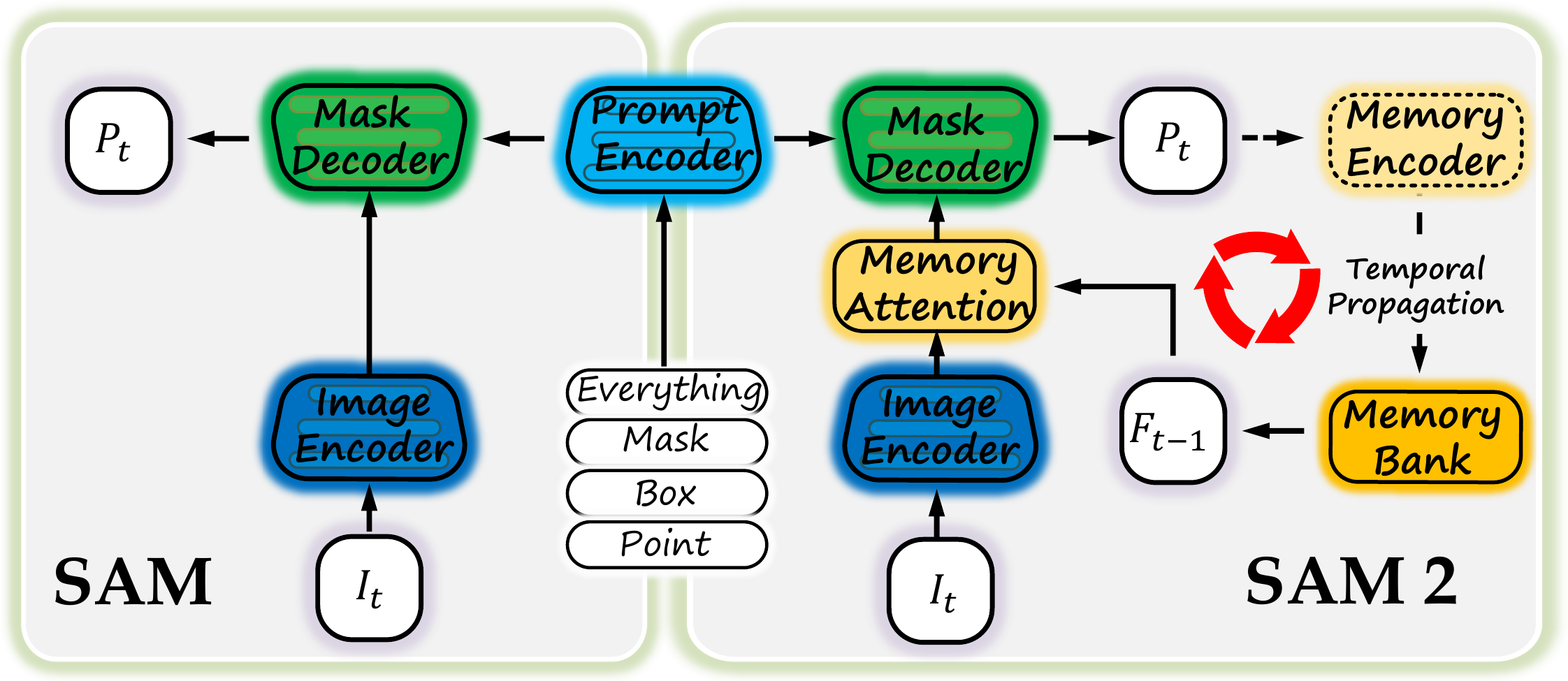}
  \caption{Architecture comparison for SAM and SAM 2.
    For the current frame $I_t$, SAM directly generates the corresponding prediction $P_t$.
    However, in SAM 2, the embedding $F_{t-1}$ from the previous prediction $P_{t-1}$ is fed into the encoding for $I_t$.}
 \label{fig:sam-1-2}
\end{figure}

The architectures of SAM and SAM 2 are delineated in Fig.~\ref{fig:sam-1-2}.
Both share a similar framework, where the image encoder extracts multi-scale features from the input image.
These features are then utilized by the mask decoder to generate prompt-specific masks, under the guidance of the information encoded by the prompt encoder. Compared with SAM, SAM 2 is enhanced with additional temporal modeling components, such as memory attention, memory encoder, and memory bank, to better leverage temporal information for video processing.

For simplicity and typicality, we uniformly evaluate the large versions of SAM and SAM 2 in all experiments on two NVIDIA V100 GPUs. 
The performance of related algorithms in various tasks is derived from the original papers, and we utilize the same evaluation tools. 
To thoroughly evaluate the capabilities of SAMs, we carefully conduct experiments with various prompt types, including
basic modes like point (\PromptPoint) and bounding box (\PromptBox) with interaction, as well as automatic segmentation (\PromptEverything) without interaction. SAM 2 also supports an additional mask type (\PromptMask).
Using these prompts, SAMs can focus on segmenting internal objects, allowing us to directly obtain the final predictions. In automatic mode (\PromptEverything), we apply an \textit{overlap filtering strategy (OFS)} based on the ground-truth mask (GT) to generate the final prediction.

\subsection{Prediction Generation}
\label{app:sec:prediction_generation}

\subsubsection{Basic Mode}
\label{app:sec:basic_mode}

\textbf{Point Prompt (\PromptPoint)}. 
To mimic interactive correction, we design an automated process that simulates successive clicking behaviors. 
At each step, the algorithm selects the pixel farthest from the correctly classified regions within either the false-positive or false-negative areas of the current prediction, and places a single corrective click at that location. 
The final prediction mask is generated through an iterative process of 
adding these corrective clicks until either the IoU exceeds 0.9 or the maximum number of clicks (6) is reached. 

\noindent\textbf{Box Prompt (\PromptBox)}. 
The process involves generating a final prediction through a series of steps:
First, all bounding boxes are obtained for connected regions within the ground truth mask (GT). Next, multiple masks are predicted using these bounding boxes. Finally, the final prediction is constructed by performing a logical OR operation across all individual masks, resulting in a union of all predicted masks.

\noindent\textbf{Mask Prompt (\PromptMask)}. 
For images, directly using the original mask as a prompt does not hold practical value.
However, for video tasks, the setup of using a mask as the prompt has been extensively explored in some context-independent concept understanding tasks like one-shot video object segmentation.
Therefore, in this work, it is only utilized in video data on specific reference frames.
And the performance of SAM 2 under this setup is tested exclusively.

\noindent\textbf{Automatic Generation (\PromptEverything)}. 
In interaction-based types including points (\PromptPoint), bounding boxes (\PromptBox), and masks (\PromptMask), we can directly obtain the final predictions through SAM or SAM 2.
However, the output in the automatic type is a segmentation map containing all local entities in the entire image, which cannot be directly used in the performance evaluation for these tasks.
Therefore, the \textit{overlap filtering strategy (OFS)} based on the ground-truth mask (GT) is employed here, retaining only those independent entities whose overlap area with the GT is greater than 90\%.
These retained entities are merged into a mask, serving as the final prediction result.

\subsubsection{In-Context Learning Mode}
\label{app:sec:incontext_lerning_mode}
Unlike SAM, SAM 2 incorporates a memory mechanism for temporal modeling, which provides it with an \textit{in-context learning (ICL) capability}~\cite{CDCU-Spider,SegGPT}. 
Instead of relying solely on prompts from the current image, SAM 2 in ICL mode integrates multiple concept exemplars to pre-encode and interpret diverse semantic representations.
To investigate this capability, we evaluate SAM 2 under the ICL setting, where we take the first 20 samples from the training dataset of each concept as the prompts, using both images and their corresponding masks as additional guidance.
This ICL approach contrasts with traditional methods constrained to single-image information, highlighting SAM 2’s advantage in leveraging exemplar-driven context understanding.

\subsubsection{SAMs for Video Segmentation}
\label{sec:perf_of_videoseg}

\noindent\textbullet~\textit{\textbf{SAM for Video Data.}}
Given that SAM is not originally designed for video data, we evaluate it 
using two distinct setups: image-based and video-based prompting.
In the image-based setup, the video is treated as a set of individual images,
where individual GT-based prompts are used to generate predictions for each frame.
In the video-based setup, we assume limited object motion and implement a propagation-based prompt strategy to assess SAM’s temporal performance without altering its architecture.
Specifically, the prompt for the current frame is automatically generated based on the prediction from the previous frame, enabling continuous prediction across the entire sequence.

\noindent\textbullet~\textit{\textbf{SAM 2 for Video Data.}}
Since objects often exhibit limited motion at the start of a video sequence, we introduce prompt information from intermediate frames. 
Specifically, we collect results under three setups: by introducing 1 frame, 3 frames, and 5 frames, referred to as ``1$\times$'', ``3$\times$'', and ``5$\times$''.
In ``1$\times$'', only the first frame is used as the object prompt.
In ``3$\times$'' and ``5$\times$'', additional frames are introduced at the $\{\frac{i}{3}\}^2_{i=1}$ and the $\{\frac{i}{5}\}^4_{i=1}$ points of the sequence, respectively.

\subsubsection{Bidirectional Inference for 3D data}
\label{app:sec:bidirectional_inference}
Some 3D medical lesion image sequences begin and end with slices containing only background, with no discernible foreground objects. 
Prompting SAM 2 on such slices may lead to unreasonable results. 
To address this, we propose a bidirectional inference strategy. 
We first traverse the sequence to identify the slice with the largest foreground region, which serves as the anchor. 
The sequence is then split into two parts: one from the start to the anchor, and the other from the anchor to the end. 
Starting from the anchor slice, SAM 2 performs bidirectional inference by propagating segmentation masks in opposite directions:
one direction traverses forward through the second subsequence toward the end of the sequence,
while the other direction traverses backward through the first subsequence toward the start of the sequence.
This dual traversal ensures that segmentation information is propagated effectively across all slices, leveraging both spatial context and temporal consistency.
After completing inference in both directions, the results are combined to form the final prediction for the entire 3D sequence.
This strategy improves segmentation accuracy by enhancing continuity across slices and mitigating the impact of slices with purely background information, ultimately enabling more reliable and consistent segmentation results for 3D medical image sequences.

\begin{figure*}[!ht]
  \centering
  \begin{minipage}{0.59\linewidth}
    \centering
       \captionof{table}{Image SOD.}
    \resizebox{\linewidth}{!}{\input{table/image_salient.tex}}
    \label{tab:image_salient}
  \end{minipage}
  \hfill
  \begin{minipage}{0.37\linewidth}
    \centering
        \captionof{table}{Image COD.}
    \resizebox{\linewidth}{!}{\input{table/image_camo.tex}}
    \label{tab:image_camo}
  \end{minipage}
  
  \begin{minipage}{0.20\linewidth}
    \centering
        \captionof{table}{Image SD.}
    \resizebox{\linewidth}{!}{\input{table/image_shadow.tex}}
    \label{tab:image_shadow}
  \end{minipage}
\hfill
  \begin{minipage}{0.24\linewidth}
    \centering
      \captionof{table}{Image TOS.}
    \resizebox{0.7\linewidth}{!}{\input{table/image_transparent.tex}}
    \label{tab:image_transparent}
  \end{minipage}
  \hfill
  \begin{minipage}{0.54\linewidth}
    \centering
      \captionof{table}{Image industrial PBD. ``\none'' indicates results unavailable as their prediction accuracy of the number of plates is zero, following the standard PBD evaluation protocol.}

    \resizebox{\linewidth}{!}{\input{table/image_battery.tex}}
    \label{tab:image_battery}
  \end{minipage}

  \begin{minipage}{0.46\linewidth}
    \centering
        \captionof{table}{Image industrial AD.}
    \resizebox{\linewidth}{!}{\input{table/image_anomaly.tex}}
    \label{tab:image_anomaly}
  \end{minipage}
  \hfill
  \begin{minipage}{0.46\linewidth}
    \centering
        \captionof{table}{Image LOS. ``\none'' denotes that each specialized model do not support segmentation of other lesion types beyond its intended specialization.}
    \resizebox{\linewidth}{!}{\input{table/image_medical.tex}}
    \label{tab:image_medical}
  \end{minipage}
\end{figure*}

\subsection{Performance of Image Segmentation}
\label{sec:perf_of_iseg}

\noindent\textbullet~\textit{\parhead{Basic Mode.}}
Tabs.~\ref{tab:image_salient},~\ref{tab:image_camo},~\ref{tab:image_shadow},~\ref{tab:image_transparent},~\ref{tab:image_battery},~\ref{tab:image_anomaly},~\ref{tab:image_medical} separately list the performance comparisons among the different specialized models and SAMs (\PromptEverything/\PromptBox/\PromptPoint) in the SOD, COD, SD, TOS, PBD, AD and LOS.
Benefiting from the ability of box prompt to filter out large amounts of background information, SAMs (\PromptBox) generally perform well across most tasks.
However, they still struggle with SD and PBD because these concepts lack clear, distinct objects and have minimal contrast with the background.
Additionally, we observe SAM 2 (\PromptPoint) and SAM 2 (\PromptEverything) are consistently weaker than their corresponding SAM variants.

\noindent\textbullet~\textit{\parhead{In-Context Learning Mode.}}
As shown in Tab.~\ref{tab:prompt20}, SAM 2 (\PromptICL) demonstrates impressive performance in segmenting these varied CD concepts.
Specifically, SAM 2 (\PromptICL) shows competitive results on TOS and SD tasks and achieves a notable lead in COD and four LOS tasks, even surpassing SAM 2 (\PromptEverything) in automatic mode. However, due to the lack of targeted training on CD concepts datasets, SAM 2 (\PromptICL) still underperforms compared to UniverSeg~\cite{UniverSeg} and Spider~\cite{CDCU-Spider}.

\begin{table*}[!t]
  \centering
  \caption{Quantitative comparison of unified models with in-context learning mode.  ``\none'' indicates that UniverSeg is a medical generalist model and does not participate in segmentation evaluations for natural scene concepts.}
  \label{tab:prompt20}
  \resizebox{\linewidth}{!}{\input{table/prompt20.tex}}
\end{table*}

\begin{table*}[!t]
  \centering
  \caption{Video LOS (Polyp Segmentation).}
  \label{tab:video_salient}
  \resizebox{0.5\linewidth}{!}{\input{table/video_polyp.tex}}
\end{table*}

\begin{figure*}[!t]
  \centering
  \begin{minipage}{0.42\linewidth}
    \centering
        \captionof{table}{Video SOD.}
    \resizebox{\linewidth}{!}{\input{table/video_salient.tex}}
    \label{tab:video_camo}
  \end{minipage}
  \hfill
  \begin{minipage}{0.26\linewidth}
    \centering
        \captionof{table}{Video COD.}
    \resizebox{\linewidth}{!}{\input{table/video_camo.tex}}
    \label{tab:video_shadow}
  \end{minipage}
  \hfill
  \begin{minipage}{0.28\linewidth}
    \centering
        \captionof{table}{Video SD.}
    \resizebox{\linewidth}{!}{\input{table/video_shadow.tex}}
    \label{tab:video_polyp}
  \end{minipage}
\end{figure*}

\begin{table*}[!t]
  \centering
  \caption{3D LOS for the whole tumor (WT), tumor core (TC) and enhancing tumor (ET), and multiple sclerosis (MS).}
  \begin{subtable}[h]{0.84\linewidth}
      \caption{BraTS2020~\cite{BraTS2020}}
    \resizebox{\linewidth}{!}{\input{table/3d_medical_brats2020.tex}}
    \label{tab:3d_medical_brats2020}
  \end{subtable}
  \hfill
  \begin{subtable}[h]{0.15\linewidth}
    \caption{ISBI2015~\cite{ISBI2015}}
    \resizebox{\linewidth}{!}{\input{table/3d_medical_isbi2015.tex}}
    \label{tab:3d_medical_isbi2015}
  \end{subtable}
  \label{tab:3d_medical}
\end{table*}

\begin{table*}[!t]
  \centering
   \caption{
    Robustness analysis under various random perturbations:
    a random perturbation (0–10\%) in the length of the shorter side of \PromptBox,
    a random displacement of up to 10 pixels in the coordinates of \PromptPoint,
    and random erosion or dilation with 5 iterations in \PromptMask.
    $\Delta$ denotes the relative performance change compared to the results with ideal prompts without perturbations.
  }
  \begin{subtable}[h]{0.74\linewidth}
    \caption{Image Data}
    \resizebox{\linewidth}{!}{\input{table/image_robustness.tex}}
    \label{tab:image_robustness}
  \end{subtable}
  \hfill
  \begin{subtable}[h]{0.24\linewidth}
    \caption{Video Data}
    \resizebox{\linewidth}{!}{\input{table/video_robustness.tex}}
    \label{tab:video_robustness}
  \end{subtable}
  \label{tab:robustness}
\end{table*}

\subsection{Performance of Video Segmentation}
\label{sec:perf_of_videoseg}

All experimental results are listed in Tabs.~\ref{tab:video_salient},~\ref{tab:video_camo},~\ref{tab:video_shadow},~\ref{tab:video_polyp}. 
We can see that SAM performs best with box prompts, followed by point prompts, and shows the lowest performance in automatic mode. This performance gap is particularly evident in challenging tasks such as COD, SD, LOS, and in complex datasets like DAVSOD$_N$ and DAVSOD$_H$ in SOD. However, with a propagation-based prompt strategy, the point form surpasses the box form and even outperforms existing domain-specific specialized models in video SOD.
For SAM 2, mask prompts yield the highest performance, followed by point and then box prompts. Both point and mask prompts show stable improvements as the number of prompts increases. In contrast, box prompts exhibit inconsistent gains, particularly on datasets like DAVSOD$_E$ and DAVSOD$_H$. Due to its built-in temporal modeling, SAM 2 demonstrates strong adaptability in video tasks, often surpassing domain-specific models with just a single prompt.
Notably, with a propagation strategy using point prompts, SAM can outperform SAM 2 with single-point prompts 
on DAVSOD$_N$ and DAVSOD$_H$ datasets.

\subsection{Performance of 3D Segmentation}
\label{sec:perf_of_3dseg}

Since some 3D medical lesion sequences consist of pure background images, we only evaluate SAM 2 using our bidirectional inference strategy. Specifically, we first traverse the 3D sequence and select the one with the largest foreground mask as the anchor. Then, the sequence is split into two parts, and SAM 2 treats each as a separate video for bidirectional inference using the shared starting frame. The combined results serve as predictions for the entire slice sequence. Each video is inferred using ``1$\times$'', ``3$\times$'', and ``5$\times$'' approaches, similar to the video setting. As shown in Tab.~\ref{tab:3d_medical}, SAM 2 (5$\times$\PromptMask) achieves excellent performance, even surpassing specialized models like 3D U-Net and DRU-Net. This demonstrates the effectiveness of bidirectional inference and multi-frame mask prompts for SAM 2.

\subsection{Prompts Robustness Analysis}
\label{sec:prompt_robustness_analysis}
Existing evaluation schemes usually use target GT to construct ideal prompts. However, this does not reflect real-world scenarios, as randomness in practical use can impact model performance. To simulate prompt randomness, we design a new evaluation scheme that introduces random perturbations to GT-based prompts. These perturbed prompts guide inference, enabling assessment of the model’s robustness.

\noindent\textbullet~\textit{\parhead{Random Perturbation for Point Mode}} (\PromptPoint).
\label{app:sec:random_point_mode}
Expanding on the multi-click strategy described in the point prompt of Sec.~\ref{app:sec:basic_mode}, we propose a perturbation method that randomly shifts the horizontal and vertical coordinates of each point by up to 10 pixels.
These random offsets introduce variations in the point placements, enabling us to evaluate the model's performance when the spatial configuration of point prompts is slightly altered.
This method provides insight into the model's robustness against coordinate distortions and uncertainties in point-based inputs.

\noindent\textbullet~\textit{\parhead{Random Perturbation for Box Mode}} (\PromptBox).
\label{app:sec:random_box_mode} 
Building upon the method described in the box prompt of Sec.~\ref{app:sec:basic_mode}, we propose a strategy to perturb bounding boxes by introducing random errors to enhance robustness against annotation noise and spatial uncertainties.
The perturbation adjusts each boundary within a maximum range of 10\% of the shorter side of the bounding box, ensuring proportional adaptability to varying object scales.
To maintain validity, the perturbed boxes are constrained to remain within the image boundaries.
This approach effectively simulates real-world imperfections, providing a more resilient foundation for model training and evaluation.

\noindent\textbullet~\textit{\parhead{Random Perturbation for Mask Mode}} (\PromptMask).
\label{app:sec:random_mask_mode} 
To assess the robustness of the model under the mask prompt strategy described in Sec.~\ref{app:sec:basic_mode}, we introduce a method that applies random morphological transformations to the input mask. Specifically, the binary mask is randomly subjected to either erosion or dilation, with the number of iterations varying up to a maximum of 5. This controlled perturbation introduces variations in the mask's boundaries and structure, enabling comprehensive evaluation of the model's ability to handle spatial distortions and inconsistencies in mask prompts.

\noindent\textbullet~\textit{\parhead{Random Perturbation on Video Data.}}
For video data, SAM uses an image-based prompting, whereas for SAM 2, we focus on the single prompt setting.

In Tab.~\ref{tab:robustness}, we present a robustness evaluation of SAM and SAM 2 on various image and video tasks. Across different datasets, perturbed prompts lead to noticeable performance fluctuations in both models. Consistent performance drops are observed for SAM and SAM 2 in the DUTS and DAVIS16 datasets. However, in other datasets, perturbed prompts occasionally help the models surpass ideal prompts. Additionally, the low standard deviation across multiple random perturbations (typically on the order of 1e-3) indicates both models' high sensitivity to perturbations. This shows a notable difference from results with ideal prompts, but with limited variation across multiple perturbations. Therefore, prompt accuracy in practical applications significantly impacts SAM segmentation performance, which is overlooked by most current studies.

\begin{figure*}[!t]
  \centering
  \includegraphics[width=0.86\linewidth]{figure/Sam_SOD_visualization.pdf}
  \caption{Visualization of salient object detection. From left to right: Image, Ground Truth, SAM (\PromptEverything), SAM (\PromptBox), SAM (\PromptPoint), SAM 2 (\PromptEverything), SAM 2 (\PromptBox), SAM 2 (\PromptPoint). }
  \label{fig:Sam_SOD_visualization}
\end{figure*}

\begin{figure*}[!t]
  \centering
  \includegraphics[width=0.86\linewidth]{figure/Sam_COD_visualization.pdf}
  \caption{Visualization of camouflaged object detection. From left to right: Image, Ground Truth, SAM (\PromptEverything), SAM (\PromptBox), SAM (\PromptPoint), SAM 2 (\PromptEverything), SAM 2 (\PromptBox), SAM 2 (\PromptPoint).}
  \label{fig:Sam_COD_visualization}
\end{figure*}

\begin{figure*}[!t]
  \centering
  \includegraphics[width=0.86\linewidth]{figure/Sam_SD_visualization.pdf}
  \caption{Visualization of shadow detection. From left to right: Image, Ground Truth, SAM (\PromptEverything), SAM (\PromptBox), SAM (\PromptPoint), SAM 2 (\PromptEverything), SAM 2 (\PromptBox), SAM 2 (\PromptPoint).}
  \label{fig:Sam_SD_visualization}
\end{figure*}

\begin{figure*}[!t]
  \centering
  \includegraphics[width=0.86\linewidth]{figure/Sam_Trans10k_visualization.pdf}
  \caption{Visualization of transparent object segmentation. From left to right: Image, Ground Truth, SAM (\PromptEverything), SAM (\PromptBox), SAM (\PromptPoint), SAM 2 (\PromptEverything), SAM 2 (\PromptBox), SAM 2 (\PromptPoint).}
  \label{fig:Sam_Trans10k_visualization}
\end{figure*}

\begin{figure*}[!t]
  \centering
  \includegraphics[width=\linewidth]{figure/Sam_PBD_visualization.pdf}
  \caption{Visualization of power battery detection. From left to right: Image, Ground Truth, SAM (\PromptEverything), SAM (\PromptBox), SAM (\PromptPoint), SAM 2 (\PromptEverything), SAM 2 (\PromptBox), SAM 2 (\PromptPoint).}
  \label{fig:Sam_PBD_visualization}
\end{figure*}

\begin{figure*}[!t]
  \centering
  \includegraphics[width=\linewidth]{figure/Sam_AD_visualization.pdf}
  \caption{Visualization of surface anomaly detection. From left to right: Image, Ground Truth, SAM (\PromptEverything), SAM (\PromptBox), SAM (\PromptPoint), SAM 2 (\PromptEverything), SAM 2 (\PromptBox), SAM 2 (\PromptPoint).}
  \label{fig:Sam_AD_visualization}
\end{figure*}

\begin{figure*}[!t]
  \centering
  \includegraphics[width=\linewidth]{figure/Sam_LOS_visualization.pdf}
  \caption{Visualization of medical lesion segmentation (Lung, Breast, Skin, Polyp). From left to right: Image, Ground Truth, SAM (\PromptEverything), SAM (\PromptBox), SAM (\PromptPoint), SAM 2 (\PromptEverything), SAM 2 (\PromptBox), SAM 2 (\PromptPoint).}
  \label{fig:Sam_LOS_visualization}
\end{figure*}

\begin{figure}[!t]
  \centering
  \includegraphics[width=\linewidth]{figure/Sam_Video_polyp_visualization.pdf}
  \caption{Visualization of video polyp segmentation. From top to bottom: Image, Ground Truth, SAM (\PromptEverything), SAM (\PromptBox), SAM (Propagated \PromptBox),  SAM (\PromptPoint), SAM (Propagated \PromptPoint),  SAM 2 (1$\times$\PromptBox), SAM 2 (3$\times$\PromptBox), SAM 2 (5$\times$\PromptBox), SAM 2 (1$\times$\PromptMask), SAM 2 (3$\times$\PromptMask), SAM 2 (5$\times$\PromptMask),  SAM 2 (1$\times$\PromptPoint),  SAM 2 (3$\times$\PromptPoint), SAM 2 (5$\times$\PromptPoint).     }
  \label{fig:Sam_Video_polyp_visualization}
\end{figure}

\begin{figure}[!t]
  \centering
  \includegraphics[width=\linewidth]{figure/Sam_VSOD_visualization.pdf}
  \caption{Visualization of video salient object detection. From top to bottom: Image, Ground Truth, SAM (\PromptEverything), SAM (\PromptBox), SAM (Propagated \PromptBox),  SAM (\PromptPoint), SAM (Propagated \PromptPoint),  SAM 2 (1$\times$\PromptBox), SAM 2 (3$\times$\PromptBox), SAM 2 (5$\times$\PromptBox), SAM 2 (1$\times$\PromptMask), SAM 2 (3$\times$\PromptMask), SAM 2 (5$\times$\PromptMask),  SAM 2 (1$\times$\PromptPoint),  SAM 2 (3$\times$\PromptPoint), SAM 2 (5$\times$\PromptPoint). }
  \label{fig:Sam_VSOD_visualization}
\end{figure}

\begin{figure}[!t]
  \centering
  \includegraphics[width=\linewidth]{figure/Sam_VCOD_visualization.pdf}
  \caption{Visualization of video camouflaged object detection. From top to bottom: Image, Ground Truth, SAM (\PromptEverything), SAM (\PromptBox), SAM (Propagated \PromptBox),  SAM (\PromptPoint), SAM (Propagated \PromptPoint),  SAM 2 (1$\times$\PromptBox), SAM 2 (3$\times$\PromptBox), SAM 2 (5$\times$\PromptBox), SAM 2 (1$\times$\PromptMask), SAM 2 (3$\times$\PromptMask), SAM 2 (5$\times$\PromptMask),  SAM 2 (1$\times$\PromptPoint),  SAM 2 (3$\times$\PromptPoint), SAM 2 (5$\times$\PromptPoint). }
  \label{fig:Sam_VCOD_visualization}
\end{figure}

\begin{figure}[!t]
  \centering
  \includegraphics[width=\linewidth]{figure/Sam_VSD_visualization.pdf}
  \caption{Visualization of video shadow detection. From top to bottom: Image, Ground Truth, SAM (\PromptEverything), SAM (\PromptBox), SAM (Propagated \PromptBox),  SAM (\PromptPoint), SAM (Propagated \PromptPoint),  SAM 2 (1$\times$\PromptBox), SAM 2 (3$\times$\PromptBox), SAM 2 (5$\times$\PromptBox), SAM 2 (1$\times$\PromptMask), SAM 2 (3$\times$\PromptMask), SAM 2 (5$\times$\PromptMask),  SAM 2 (1$\times$\PromptPoint),  SAM 2 (3$\times$\PromptPoint), SAM 2 (5$\times$\PromptPoint). }
  \label{fig:Sam_VSD_visualization}
\end{figure}

\begin{figure}[!t]
  \centering
  \includegraphics[width=0.96\linewidth]{figure/Sam_ISBI3D_visualization.pdf}
  \caption{ Visualization of 3D brain tumor segmentation. From top to bottom: Image, Ground Truth, SAM 2 (1$\times$\PromptBox), SAM 2 (3$\times$\PromptBox), SAM 2 (5$\times$\PromptBox), SAM 2 (1$\times$\PromptMask), SAM 2 (3$\times$\PromptMask), SAM 2 (5$\times$\PromptMask),  SAM 2 (1$\times$\PromptPoint),  SAM 2 (3$\times$\PromptPoint), SAM 2 (5$\times$\PromptPoint).}
  \label{fig:Sam_ISBI3D_visualization}
\end{figure}

\subsection{Performance Summary}
\label{sec:Performance_Summary}

Through the aforementioned comprehensive evaluation, the performance of SAMs on context-dependent concept segmentation can be summarized as follows:
\noindent\textit{\textbf{\uppercase\expandafter{\romannumeral1)}}} The box prompt is generally the most advantageous type of prompt for SAMs.
\noindent\textit{\textbf{\uppercase\expandafter{\romannumeral2)}}} SAM 2 does not always outperform SAM and performs worse on tasks involving everything and point prompts. We hypothesize that SAM 2's temporal memory and dynamic prompt attention, while beneficial in video and 3D tasks, may introduce unfavorable inductive bias in static single-frame settings. Without temporal context, the memory module may add noise or redundancy, and dynamic prompt attention may become overly adaptive, leading to unstable predictions under sparse prompts.
\noindent\textit{\textbf{\uppercase\expandafter{\romannumeral3)}}} SAM 2 demonstrates potential for in-context learning (ICL) predictions, but further exploration is needed.
\noindent\textit{\textbf{\uppercase\expandafter{\romannumeral4)}}} In video segmentation, SAM completes one-shot video object segmentation by propagating a point prompt from the first frame, showing that an image-trained model can handle video tasks effectively, while SAM 2 performs even better and surpasses specialized models.
\noindent\textit{\textbf{\uppercase\expandafter{\romannumeral5)}}} In 3D medical lesion segmentation, the proposed bidirectional inference strategy and multi-frame mask prompts help SAM 2 achieve excellent performance, even surpassing specialized models.\noindent\textit{\textbf{\uppercase\expandafter{\romannumeral6)}}} SAMs perform poorly on non-material or extremely small target concepts, such as shadows or power battery plate endpoints.\noindent\textit{\textbf{\uppercase\expandafter{\romannumeral7)}}} SAMs are highly sensitive to the accuracy of prompts.

We further analyze correlations among the evaluated CD concepts through universal challenges and scenario-specific divergences.

\noindent\textbf{Universal CD Concept Challenges.}  
All 11 CD concepts share three core dependencies that drive consistent SAM behaviors.  \textit{{\uppercase\expandafter{\romannumeral1}) Context dependency as performance bottleneck.}} CD concepts inherently rely on global and local context (e.g., camouflage requires background contrast, lesions need tissue reference). This amplifies prompt design sensitivity: box prompts dominate in 10/11 concepts (Tab.~\ref{tab:image_salient}–Tab.~\ref{tab:image_medical}), while SAMs degrade sharply when targets blend with backgrounds (e.g., BER $>$ 0.15 in SD task; PN-ACC $<$ 0.2 in PBD task). Prompt robustness tests (Tab.~\ref{tab:robustness}) further confirm universal sensitivity to perturbation.  \textit{{\uppercase\expandafter{\romannumeral2}) Temporal/sequential context benefits SAM~2.}} In video and 3D tasks (5/11 concepts), SAM~2’s memory attention yields substantial gains. Five-frame mask prompts improve COD MAE by 60\% over SAM, while bidirectional inference raises brain tumor Dice by 16\% compared to DRUNet (Tab.~\ref{tab:3d_medical_isbi2015}). Gains consistently scale with frame count (Tab.~\ref{tab:video_polyp}-Tab.~\ref{tab:3d_medical}), highlighting temporal modeling as a universal advantage.

\noindent\textbf{Scenario-Specific Divergences.}  
Performance variance arises from scene modality and concept attributes.  
\textit{{\uppercase\expandafter{\romannumeral1}) Scene modality dictates adaptation.}} In natural image tasks (SOD/COD/SD/TOS), box prompts dominate (F$^{\omega}_{\beta}$ $>$ 0.85), and SAM~2 excels in videos. In medical image tasks (LOS), box prompts are crucial for small lesions, with 3D data requiring SAM~2’s bidirectional inference. In industrial tasks (PBD/AD), prompts fail for PBD (PN-ACC $<$ 0.2), while AD depends on mask prompts for texture cues.  
\textit{{\uppercase\expandafter{\romannumeral2})  Difficulty tiering.}} Comparing SAMs with specialist models, we categorize CD tasks into: High difficulty (SD, PBD, TOS, AD) with worst results (e.g., BER $>$ 0.15 in SD task); Medium difficulty (COD, lung/breast/brain lesions) where box prompts rescue performance; and Low difficulty (SOD, skin/polyps lesions) with clear contrast, where SAMs approach SOTA (e.g., F$^{\omega}_{\beta}$ $>$ 0.92 on DUTS in SOD tasks, Dice $>$ 0.91 in polyp segmentation) performance.

\subsection{Prediction Visualization}
\label{app:sec:Visualization}
Fig.~\ref{fig:Sam_SOD_visualization},~\ref{fig:Sam_COD_visualization},~\ref{fig:Sam_SD_visualization},~\ref{fig:Sam_Trans10k_visualization},~\ref{fig:Sam_PBD_visualization},~\ref{fig:Sam_AD_visualization},~\ref{fig:Sam_LOS_visualization},~\ref{fig:Sam_Video_polyp_visualization},~\ref{fig:Sam_VSOD_visualization},~\ref{fig:Sam_VCOD_visualization},~\ref{fig:Sam_VSD_visualization},~\ref{fig:Sam_ISBI3D_visualization} show qualitative comparisons between SAM and SAM 2 across diverse CD concepts. These results complement quantitative findings and reveal several insights.

\textit{\textbf{{\uppercase\expandafter{\romannumeral1})}} \textbf{Effectiveness of box prompts.}}
In static image tasks such as SOD (Fig.~\ref{fig:Sam_SOD_visualization}), COD (Fig.~\ref{fig:Sam_COD_visualization}), and LOS (Fig.~\ref{fig:Sam_LOS_visualization}), both models produce coherent masks under box prompts, whereas point or everything prompts often yield incomplete or noisy regions.  
Box prompts filter distracting background and focus inference on the foreground, stabilizing boundaries when foreground and background share textures.  
This agrees with Tab.~\ref{tab:image_salient}–\ref{tab:image_medical}, where box prompts dominate in 10/11 concepts and deliver the best Dice/F$^{\omega}_{\beta}$/MAE overall.  
In COD specifically, box prompts not only raise Dice but also reduce holes and spurious fragments relative to point prompts, indicating the value of spatially constrained guidance.

\textit{\textbf{{\uppercase\expandafter{\romannumeral2})}} \textbf{Failure cases in small or non-material targets.}}  
For SD (Fig.~\ref{fig:Sam_SD_visualization}), TOS (Fig.~\ref{fig:Sam_Trans10k_visualization}), and PBD (Fig.~\ref{fig:Sam_PBD_visualization}), both SAM and SAM 2 struggle to capture fine structures or non-material boundaries, often over-segmenting or missing targets completely.  
These concepts require the model to capture subtle contextual cues. Shadows are defined by illumination rather than material boundaries, while transparent objects blend with the background.  
Similarly, power battery endpoints appear as micro-scale irregularities requiring near-pixel precision.

\textit{\textbf{{\uppercase\expandafter{\romannumeral3})}} \textbf{Advantages of SAM~2 in temporal and 3D settings.}}
In video tasks such as VSOD, VCOD, and VSD, SAM~2 maintains temporal consistency and preserves object boundaries more effectively than SAM, especially with multi-frame prompts.
Whereas SAM’s predictions fluctuate frame by frame, SAM~2 leverages memory attention to stabilize object identity and reduce flicker, yielding smoother trajectories and crisper edges.
In 3D brain tumor segmentation (Fig.~\ref{fig:Sam_ISBI3D_visualization}), the bidirectional inference strategy ensures continuity across slices, allowing SAM~2 to capture volumetric structures that SAM underestimates or fragments.

\textit{\textbf{{\uppercase\expandafter{\romannumeral4})}} \textbf{Cross-domain robustness gaps.}}  
In anomaly detection (Fig.~\ref{fig:Sam_AD_visualization}) and medical lesions (Fig.~\ref{fig:Sam_LOS_visualization}), predictions reveal high sensitivity to prompt perturbations and cluttered backgrounds.  
When masks are eroded or dilated, outputs often collapse, either shrinking excessively or bleeding into irrelevant regions, which highlights the brittle reliance on precise prompts.  
Although SAM~2 typically yields smoother and more aligned masks, both models degrade in low-contrast or ambiguous contexts, consistent with Tab.~\ref{tab:robustness}.  
These behaviors indicate limited robustness in real-world deployments without prompt quality control or additional context modeling.

\clearpage

\section{Outlook for SAM 3}
\label{sec:outlook}

In this section, we analyze the characteristics of current popular unified segmentation models, including
SAM~\cite{SAM},
SAM 2~\cite{SAM2},
UniverSeg~\cite{UniverSeg},
SegGPT~\cite{SegGPT},
and Spider~\cite{CDCU-Spider},
across the following aspects:
architecture,
prompt types,
prompt-target interaction,
training data and strategies.
In this way, we can provide a meaningful outlook for SAM 3.

\begin{figure}[!t]
  \centering
  \includegraphics[width=\linewidth]{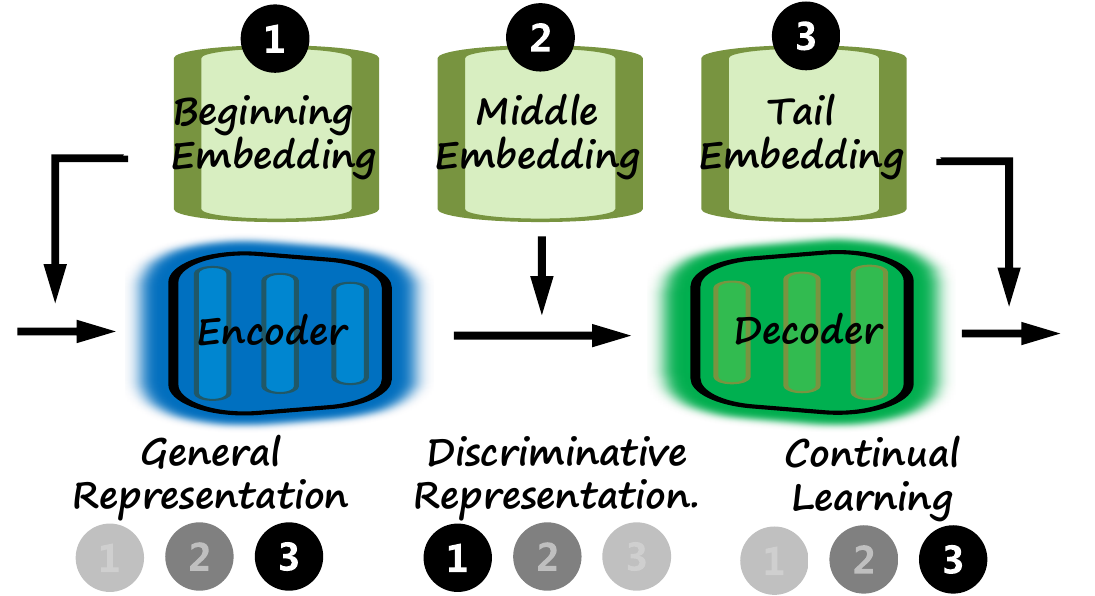}
   \caption{Architecture with three different embedding positions.}
    \label{fig:Architecture_comparison}
\end{figure}

\noindent\textbullet~\textit{\textbf{Architecture.}}
Unified segmentation models typically utilize a straightforward encoder-decoder framework without elaborate modules.
They segment prompt-defined concepts through interactions between prompt and target features.
As shown in Fig.~\ref{fig:Architecture_comparison}, these models employ different strategies for embedding prompts:
UniverSeg and SegGPT use beginning embedding,
SAM and SAM 2 use middle embedding,
and Spider uses tail embedding.
Key capabilities for a strong segmentation model include
representing general concepts,
distinguishing different features,
and enabling continuous learning.
The position of prompt embedding significantly impacts these capabilities.
For instance, beginning embedding tightly integrates the prompt with the concept from the outset, enhancing discriminative representation by focusing on concept distinctions.
However, it reduces general representation capability and complicates continual learning, requiring fine-tuning of the entire network for new concepts.
Conversely, tail embedding offers a different strategy, while middle embedding provides a more balanced solution.
Future advancements in prompt information propagation could address tail embedding's weaknesses in discriminative representation, making it more competitive.

\begin{figure}[!t]
  \centering
  \includegraphics[width=\linewidth]{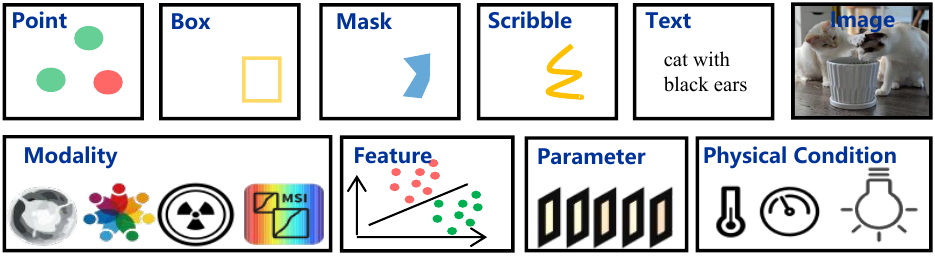}
   \caption{Diverse prompt types.}
    \label{fig:prompt_types}
\end{figure}

\noindent\textbullet~\textit{\textbf{Prompt Types.}}
Fig.~\ref{fig:prompt_types} illustrates various prompt types used or yet to be utilized in unified models.
Currently, popular types include point, box, mask, text, and image prompts.
To improve segmentation across diverse scenarios, exploring new prompt types is key.
Potential directions include:
\noindent\textit{\textbf{\uppercase\expandafter{\romannumeral1)}}}
Modalities like depth maps, infrared images, multispectral images, and X-rays can provide valuable context beyond traditional RGB.
These data types help models better understand scene and object structures, especially in medical imaging and industrial inspection.
\noindent\textit{\textbf{\uppercase\expandafter{\romannumeral2)}}}
Predefined features or attributes, such as high-dimensional vectors or task-specific attributes, can guide segmentation, particularly in domain-specific tasks.
For example, in industrial battery detection, feature prompts representing pristine electrodes can help identify anomalies more accurately.
\noindent\textit{\textbf{\uppercase\expandafter{\romannumeral3)}}}
These prompts dynamically adjust the model's parameters, similar to learnable prompts but focused on optimizing weights and structure.
Existing image restoration methods have shown that learnable parameter cues can capture unknown degradation types, improving tasks like denoising, deblurring, and restoration across various domains.
\noindent\textit{\textbf{\uppercase\expandafter{\romannumeral4)}}}
In sensor-based scenarios, prompts can use real-time environmental data, such as temperature or motion, to guide system behavior.
For instance, wearable medical devices can personalize responses based on individual physiological data, while industrial systems can adapt based on specific environmental conditions for optimized user experience.

Moreover, the current isolated prompt strategy often lacks sufficient context. In real-world applications, multiple prompt types can be gathered simultaneously.
Developing a unified prompt embedding mechanism to integrate these types could create a truly unified structure, enhancing segmentation capabilities across diverse scenarios.

\begin{figure*}[!t]
	\centering
	\includegraphics[width=0.94\linewidth]{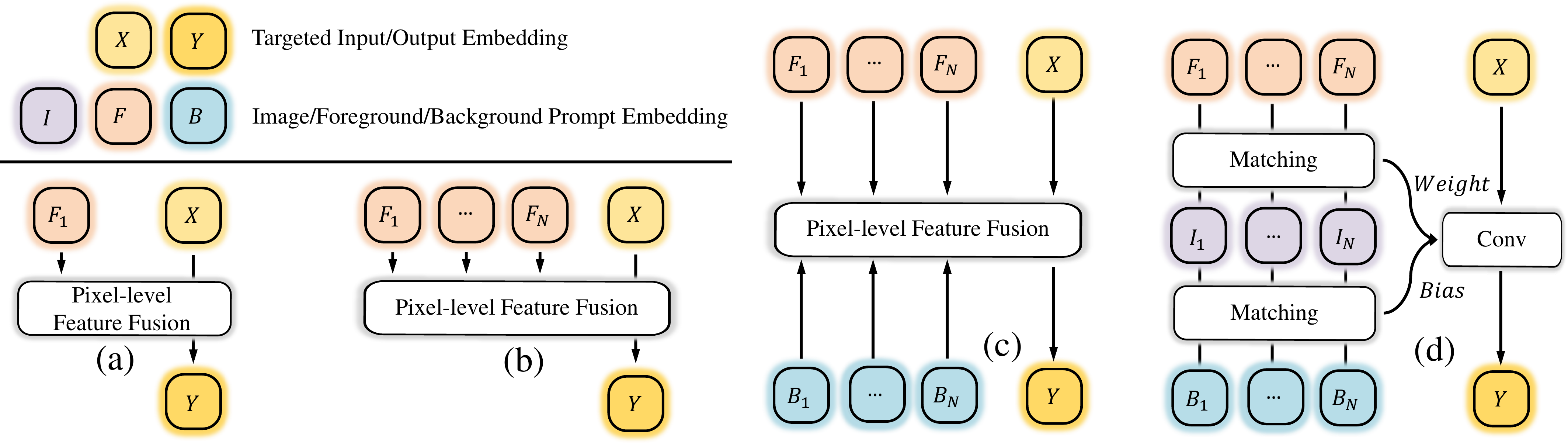}
	\caption{Four types of feature interaction between visual prompts and current target input.
	}
	\label{fig:Prompt_Interaction}
\end{figure*}

\noindent\textbullet~\textit{\textbf{Prompt-Target Interaction.}}
The effective interaction between target features and prompt features is the key to drive a unified model to distinguish different concepts based on limited prompts.
SegGPT uses a single foreground prompt embedding for pixel-level feature fusion (Fig.~\ref{fig:Prompt_Interaction}(a)).
UniverSeg employs group foreground prompt embedding for the same purpose (Fig.~\ref{fig:Prompt_Interaction}(b)), while SAM 2 (\PromptPoint) uses both foreground and background group prompt embeddings (Fig.~\ref{fig:Prompt_Interaction}(c)).
Spider condenses high-level image-foreground and image-background matching knowledge to generate a concept filter that facilitates feature interaction (Fig.~\ref{fig:Prompt_Interaction}(d)). 
Group prompt embeddings are gaining popularity as they explicitly enhance prompt information.
Pixel-level fusion excels in perceiving consistent target appearances, making it effective for context-independent (CI) concepts.
However, appearance variations may cause ambiguity, limiting its effectiveness for context-dependent (CD) concepts.
In contrast, high-level concept filtering relies on abstract information, allowing Spider to excel in CD concepts.
While Spider can handle CI tasks, it tends to focus more on object localization, overlooking appearance details.
Future work could combine both interaction forms to improve segmentation for both CI and CD concepts.

\noindent\textbullet~\textit{\textbf{Training Data and Strategies.}} 
Most unified models aim to obtain strong representations from large datasets to improve generalization across various concepts.
However, there is no benchmark dataset specifically for unified segmentation models.
Spider is trained exclusively on datasets with context-dependent (CD) concepts, while others use context-independent (CI) datasets.
Integrating both CD and CI concepts with a self-training strategy, similar to SAM, could create a large-scale CI-CD joint benchmark by annotating different concepts for each image, benefiting segmentation and enhancing model discriminative power.
Additionally, concept-balanced training is essential.
SegGPT assigns different sampling weights to balance concepts from the data scale perspective,
while SAM simulates interactive segmentation by iteratively refining masks with initial prompts.
In contrast, UniverSeg focuses on enhancing data sample diversity.
Spider considers concept balance in propagation, but its resource limitations prevent simultaneous training on multiple concepts.
A potential future direction could be adjusting learning rates and update directions based on concept performance, inspired by optimizers like SGD and Adam~\cite{Adam}, to improve convergence and balance across concepts.
Moreover, joint multi-modal training may provide a natural solution to bridge CI and CD gaps, enabling shared features while preserving domain-specific nuances.

\section{Limitations}
\label{app:sec:Limitations}
We outline improvement suggestions in Sec.~\ref{sec:outlook}, summarizing characteristics of unified segmentation models. While full implementation would require dedicated research (potentially spanning multiple studies), such engineering efforts including novel module architectures and large-scale dataset curation, exceed the scope of this evaluation-focused work. Evaluating the feasibility of these outlook suggestions and implementing them will require collaborative efforts from the broader research community. 
Current evaluations are also constrained by dataset diversity and annotation quality, which may under-represent real deployment challenges.

\section{Conclusion}
\label{sec:conclusion}

This paper provides a comprehensive evaluation of SAMs (SAM and SAM 2) in segmenting context-dependent (CD) concepts across 11 categories with 2D, 3D, and video data in natural, medical, and industrial scenes. First, we establish a unified inference framework for SAM and SAM 2 to assess prompt types, strategies, and robustness.
Next, we conduct extensive experiments on SAMs across different concepts in image, video, and 3D data, during which we also demonstrate the effectiveness of the proposed propagation-based prompt strategy, bidirectional inference strategy, and in context learning-based inference mode. This enables us to discuss the strengths and limitations of SAMs in segmenting CD concepts.
Finally, we summarize the characteristics of various unified segmentation models and provide suggestions for improvement. Based on these results and insights, we believe this work will establish a baseline for CD concept segmentation and encourage further enhancement of SAM 2 in anticipation of SAM 3.




\bibliographystyle{abbrv}
\bibliography{main}

\end{document}

%% file: table/survey_comparison.tex
\begin{tabular}{lllllll}
  \toprule[2pt]
  \textbf{Method}                             & \textbf{Scene} & \textbf{CD Concept } & \textbf{Datasets}                                                                                                & \textbf{Modality}      & \multicolumn{2}{c}{\textbf{Prompt Modes}}                      \\
  \midrule[1pt]
  Ji~\textit{et al.}~\cite{Ji_SAM_report1}    & Natural        & Camouflaged Object   & CAMO~\cite{CAMO}, COD10K~\cite{SINet_COD}, NC4K~\cite{Rank-Net_COD}, CHAMELEON~\cite{CHAMELEON}                             & RGB Image              & \PromptEverything                                        &     \\
  \midrule[1pt]
  Ji~\textit{et al.}~\cite{Ji_SAM_report2}    & Natural        & Salient Object       & DUTS~\cite{DUTS}, COME15K-Diff~\cite{CMINet_RGBDSOD}, VT1000~\cite{VT1000-SDGL}, DIS-TE4~\cite{DIS}                                                                                                     & RGB Image              & \PromptEverything                                        &     \\
  \cmidrule[0.5pt](r){3-7}                    &                & Camouflaged Object   &   COD10K~\cite{SINet_COD}, CDS2K~\cite{CDS2K}                                                                                                               & RGB Image              &     \PromptEverything                                                      &     \\
  \cmidrule[1pt](r){3-7}                      &                & Shadow Object        & SBU~\cite{SBU}                                                                                                                 &   RGB Image                     &         \PromptEverything                                                  &     \\
  \cmidrule[0.5pt](r){2-7}                    & Medical        & Polyp Lesion         &  CVC-ColonDB~\cite{CVC-ColonDB}                                                                    &     Endoscopy                     &       \PromptEverything                                                    &     \\
  \midrule[1pt]
  Zhou~\textit{et al.}~\cite{Zhou_SAM_report} & Medical        & Polyp Lesion         &   Kvasir~\cite{Kvasir}, ETIS~\cite{ETIS}, CVC-ClinicDB~\cite{CVC-ClinicDB}, CVC-ColonDB~\cite{CVC-ColonDB}, Endoscene~\cite{Endoscene}                                                                                                                &       Endoscopy       & \PromptEverything                                        &     \\
  \midrule[1pt]
  Tang and Li~\cite{Tang_SAM_report}          & Natural        & Camouflaged Object   &     CAMO~\cite{CAMO}, COD10K~\cite{SINet_COD}, NC4K~\cite{Rank-Net_COD}                                                                                                             & RGB Image              & \PromptEverything, \PromptBox                                             &     \\
                                              &                &                      &              MoCA-Mask~\cite{VCOD-MoCA}                                                                                                   & RGB Video              & \PromptPoint                                               &     \\
  \midrule[1pt]
  Lian and Li~\cite{Lian_SAM_report}          & Natural        & Salient Object       &   USIS10K~\cite{USIS10K}                                                                                                               & RGB Image               & \PromptEverything, \PromptBox, \PromptPoint              &     \\
  \midrule[1pt]
  Chen~\textit{et al.}~\cite{Chen_SAM_report} & Natural        & Camouflaged Object   &  CHAMELEON~\cite{CHAMELEON}, CAMO~\cite{CAMO}, COD10K~\cite{SINet_COD}                                                                                                                & RGB Image              & \PromptEverything                                        &     \\
     \cmidrule[0.5pt](r){3-7}                                             &              & Shadow Object        &      ISTD~\cite{ISTD}                                                                                                              &RGB Image                        &      \PromptEverything                                                    &     \\
    \cmidrule[0.5pt](r){2-7}                    & Medical        & Polyp Lesion         &  Kvasir~\cite{Kvasir}                                                                    &     Endoscopy                     &       \PromptEverything                                                    &     \\
  \midrule[1pt]
  Ours                                        & Natural        & Salient Object       & DUTS~\cite{DUTS}, ECSSD~\cite{ECSSD}, DUT-OMRON~\cite{DUT-OMRON}, HKU-IS~\cite{HKU-IS}, PASCAL-S~\cite{PASCAL-S} & RGB Image              & \PromptEverything, \PromptBox, \PromptPoint, \PromptICL  & +PR \\
                                              &                &                      & DAVIS-16~\cite{DAVIS16}, DAVSOD~\cite{DAVSOD}                                                                                 & RGB Video              & \PromptEverything, \PromptBox, \PromptPoint, \PromptMask & +PR \\
  \cmidrule[0.5pt](r){3-7}
                                              &                & Camouflaged Object   & CAMO~\cite{CAMO}, COD10K~\cite{SINet_COD}, NC4K~\cite{Rank-Net_COD}                                                         & RGB Image              & \PromptEverything, \PromptBox, \PromptPoint, \PromptICL  &     \\
                                              &                &                      & CAD~\cite{VCOD-CAD}, MoCA-Mask~\cite{VCOD-MoCA-Mask}                                                             & RGB Video              & \PromptEverything, \PromptBox, \PromptPoint, \PromptMask &     \\
  \cmidrule[0.5pt](r){3-7}
                                              &                & Shadow Object        & SBU~\cite{SBU}, ISTD~\cite{ISTD}                                                                                        & RGB Image              & \PromptEverything, \PromptBox, \PromptPoint, \PromptICL  &     \\
                                              &                &                      & VISAD-DS~\cite{VISAD-DS}, VISAD-MOS~\cite{VISAD-DS}                                                                              & RGB Video              & \PromptEverything, \PromptBox, \PromptPoint, \PromptMask &     \\
  \cmidrule[0.5pt](r){3-7}
                                              &                & Transparent Object   & Trans10K~\cite{TransLab_Transparent}                                                                                                 & RGB Image              & \PromptEverything, \PromptBox, \PromptPoint, \PromptICL  &     \\
  \cmidrule[1pt](r){2-7}
                                              & Medical        & Polyp Lesion         & Kvasir~\cite{Kvasir}, ETIS~\cite{ETIS}, CVC-ClinicDB~\cite{CVC-ClinicDB}, CVC-ColonDB~\cite{CVC-ColonDB}, Endoscene~\cite{Endoscene}                       & Endoscopy Image              & \PromptEverything, \PromptBox, \PromptPoint, \PromptICL  &     \\
                                              &                &                      & CVC-612-T~\cite{PNS-Net_Polyp}, CVC-612-V~\cite{PNS-Net_Polyp}, CVC-300-TV~\cite{PNS-Net_Polyp}                                      & Endoscopy Video              & \PromptEverything, \PromptBox, \PromptPoint, \PromptMask &     \\
  \cmidrule[0.5pt](r){3-7}
                                              &                & Skin Lesion      & ISIC-2018~\cite{ISIC18}                                                                                                & Dermoscopy          & \PromptEverything, \PromptBox, \PromptPoint, \PromptICL  &     \\
  \cmidrule[0.5pt](r){3-7}
                                              &                & Lung Infection       & COVID-19 CT~\cite{Inf-Net}                                                                                              & CT                     & \PromptEverything, \PromptBox, \PromptPoint, \PromptICL  &     \\
  \cmidrule[0.5pt](r){3-7}
                                              &                & Brain Tumor          & BraTS2020~\cite{BraTS2020}, ISBI2015~\cite{ISBI2015}                                                                            & MRI (T1/T2/T1ce/Flair) & \PromptEverything, \PromptBox, \PromptPoint              &  \\
  \cmidrule[0.5pt](r){3-7}
                                              &                & Breast Lesion        & BUSI~\cite{BUSI}                                                                                                     & Ultrasound             & \PromptEverything, \PromptBox, \PromptPoint, \PromptICL  &     \\
  \cmidrule[1pt](r){2-7}
                                              & Industrial     & Power Battery Plate      & PBD~\cite{AI4Industry-PBD} (Regular/Difficult/Tough)                                                             & X-ray                  & \PromptEverything, \PromptBox, \PromptPoint              &     \\
  \cmidrule[0.5pt](r){3-7}
                                              &                & Surface Anomaly              & MVTec-AD~\cite{MVTec-AD}, VisA~\cite{VisA}, BTAD~\cite{BTAD}                                                                     & RGB Image             & \PromptEverything, \PromptBox, \PromptPoint              & +PR \\
                                         
  \bottomrule[2pt]
\end{tabular}

%% file: table/dataset_link.tex
\begin{tabular}{l|c|c}
\toprule[2pt]
\textbf{Task} & \textbf{Dataset Name} & \textbf{Link}  \\
  \midrule[1pt]
Image Salient Object Detection&DUTS&\url{http://saliencydetection.net/duts}\\
Image Salient Object Detection&ECSSD&\url{https://www.cse.cuhk.edu.hk/~leojia/projects/hsaliency/dataset.html}\\
Image Salient Object Detection&DUT-OMRON&\url{http://saliencydetection.net/dut-omron}\\
Image Salient Object Detection&HKU-IS&\url{https://arxiv.org/abs/1503.08663}\\
Image Salient Object Detection&PASCAL-S&\url{https://ccvl.jhu.edu/datasets}\\
Video Salient Object Detection&DAVIS16&\url{https://davischallenge.org/davis2016/code.html}\\
Video Salient Object Detection&DAVSOD&\url{https://github.com/DengPingFan/DAVSOD}\\
Image Camouflaged Object Detection&CAMO&\url{https://sites.google.com/view/ltnghia/research/camo}\\
Image Camouflaged Object Detection&COD10K&\url{https://github.com/DengPingFan/SINet}\\
Image Camouflaged Object Detection&NC4K&\url{https://github.com/JingZhang617/COD-Rank-Localize-and-Segment}\\
Video Camouflaged Object Detection&CAD&\url{http://vis-www.cs.umass.edu/motionSegmentation}\\
Video Camouflaged Object Detection&MoCA-Mask&\url{https://github.com/XuelianCheng/SLT-Net}\\
Image Shadow Detection&SBU&\url{https://www3.cs.stonybrook.edu/\~cvl/projects/shadow\_noisy\_label}\\
Image Shadow Detection&ISTD&\url{https://github.com/DeepInsight-PCALab/ST-CGAN}\\
Video Shadow Detection&VISAD&\url{https://github.com/yihong-97/STICT}\\
Transparent Object Segmentation&Trans10K&\url{https://xieenze.github.io/projects/TransLAB/TransLAB.html}\\
Image Polyp Lesion  Segmentation&Kvasir&\url{https://datasets.simula.no/kvasir-seg}\\
Image Polyp Lesion  Segmentation&ETIS&\url{https://polyp.grand-challenge.org/ETISLarib}\\
Image Polyp Lesion  Segmentation&CVC-ClinicDB&\url{https://polyp.grand-challenge.org/CVCClinicDB}\\
Image Polyp Lesion  Segmentation&CVC-ColonDB&\url{http://vi.cvc.uab.es/colon-qa/cvccolondb}\\
Image Polyp Lesion  Segmentation&Endoscene&\url{https://arxiv.org/abs/1612.00799}\\
Video Polyp Lesion  Segmentation&Video Polyp&\url{https://github.com/GewelsJI/PNS-Net}\\
Skin Lesion  Segmentation&ISIC-2018&\url{https://challenge.isic-archive.com/landing/2018}\\
COVID-19 Lung Infection Segmentation&COVID-19 CT&\url{https://github.com/DengPingFan/Inf-Net}\\
Brain Tumor Segmentation&BraTS2020&\url{https://www.med.upenn.edu/cbica/brats2020}\\
Brain Tumor Segmentation&ISBI2015&\url{https://smart-stats-tools.org/lesion-challenge-2015}\\
Breast Lesion Segmentation&BUSI&\url{https://scholar.cu.edu.eg/?q=afahmy/pages/dataset}\\
Power Battery Detection&X-ray PBD&\url{https://github.com/Xiaoqi-Zhao-DLUT/X-ray-PBD}\\
Surface Anomaly Detection&MVTec-AD&\url{https://www.mvtec.com/company/research/datasets/mvtec}\\
Surface Anomaly Detection&VisA&\url{https://github.com/amazon-science/spot-diff}\\
Surface Anomaly Detection&BTAD&\url{https://ieeexplore.ieee.org/abstract/document/9576231}\\

\bottomrule[2pt]
\end{tabular}

%% file: table/image_salient.tex
\begin{tabular}{l|cc|cc|cc|cc|cc}
  \toprule[2pt]
                                 &
  \multicolumn{2}{c|}{DUTS~\cite{DUTS}}      &
  \multicolumn{2}{c|}{PASCAL-S~\cite{PASCAL-S}}  &
  \multicolumn{2}{c|}{DUT-OMRON~\cite{DUT-OMRON}} &
  \multicolumn{2}{c|}{ECSSD~\cite{ECSSD}}     &
  \multicolumn{2}{c}{HKU-IS~\cite{HKU-IS}}                                                                                     \\
                                 & F$^{\omega}_{\beta}\uparrow$   & Sm$\uparrow$    & F$^{\omega}_{\beta}\uparrow$   & Sm$\uparrow$    & F$^{\omega}_{\beta}\uparrow$   & Sm$\uparrow$    & F$^{\omega}_{\beta}\uparrow$   & Sm$\uparrow$    & F$^{\omega}_{\beta}\uparrow$   & Sm$\uparrow$    \\
  \midrule[1pt]
  EDN~\cite{SOD-EDN}                            & 0.844 & 0.892 & 0.827 & 0.865 & 0.770 & 0.850 & 0.918 & 0.927 & 0.908 & 0.924 \\
  MENet~\cite{SOD-MENet}                          & 0.876 & 0.897 & 0.848 & 0.861 & 0.775 & 0.843 & 0.924 & 0.922 & 0.922 & 0.921 \\
  \midrule[1pt]
  SAM (\PromptEverything)        & 0.884 & 0.896 & 0.719 & 0.784 & 0.898 & 0.906 & 0.957 & 0.942 & 0.939 & 0.930 \\
  SAM (\PromptBox)               & 0.920 & 0.910 & 0.750 & 0.801 & 0.933 & 0.924 & 0.950 & 0.933 & 0.923 & 0.908 \\
  SAM (\PromptPoint)             & 0.886 & 0.888 & 0.760 & 0.803 & 0.931 & 0.929 & 0.964 & 0.949 & 0.927 & 0.919 \\
  \midrule[1pt]
  SAM 2 (\PromptEverything)      & 0.449 & 0.661 & 0.514 & 0.668 & 0.545 & 0.709 & 0.719 & 0.795 & 0.712 & 0.790 \\
  SAM 2 (\PromptBox)             & 0.929 & 0.921 & 0.752 & 0.801 & 0.941 & 0.932 & 0.958 & 0.941 & 0.928 & 0.914 \\
  SAM 2 (\PromptPoint)           & 0.807 & 0.815 & 0.634 & 0.688 & 0.772 & 0.777 & 0.777 & 0.766 & 0.759 & 0.756 \\
  \bottomrule[2pt]
\end{tabular}

%% file: table/image_camo.tex
\begin{tabular}{l|cc|cc|cc}
  \toprule[2pt]
                                              &
  \multicolumn{2}{c}{COD10K~\cite{SINet_COD}} &
  \multicolumn{2}{|c}{CAMO~\cite{CAMO}}       &
  \multicolumn{2}{|c}{NC4K~\cite{Rank-Net_COD}}                                                                                                                                         \\
                                              & F$^{\omega}_{\beta}\uparrow$ & Sm$\uparrow$ & F$^{\omega}_{\beta}\uparrow$ & Sm$\uparrow$ & F$^{\omega}_{\beta}\uparrow$ & Sm$\uparrow$ \\
  \midrule[1pt]
  SARNet~\cite{COD-SARNet}                        & 0.820                        & 0.885        & 0.844                        & 0.874        & 0.851                        & 0.889        \\
  ZoomNext~\cite{Zoomnext}                    & 0.838                        & 0.898        & 0.859                        & 0.888        & 0.865                        & 0.900        \\
  \midrule[1pt]
  SAM (\PromptEverything)                     & 0.694                        & 0.786        & 0.631                        & 0.707        & 0.698                        & 0.773        \\
  SAM (\PromptBox)                            & 0.863                        & 0.882        & 0.853                        & 0.854        & 0.878                        & 0.885        \\
  SAM (\PromptPoint)                          & 0.823                        & 0.868        & 0.843                        & 0.862        & 0.846                        & 0.876        \\
  \midrule[1pt]
  SAM 2 (\PromptEverything)                   & 0.260                        & 0.587        & 0.170                        & 0.493        & 0.237                        & 0.550        \\
  SAM 2 (\PromptBox)                          & 0.902                        & 0.911        & 0.891                        & 0.891        & 0.920                        & 0.918        \\
  SAM 2 (\PromptPoint)                        & 0.864                        & 0.868        & 0.771                        & 0.784        & 0.854                        & 0.851        \\
  \bottomrule[2pt]
\end{tabular}

%% file: table/image_shadow.tex
\begin{tabular}{l|c|c}
  \toprule[2pt]
                            & ISTD~\cite{ISTD}             & SBU~\cite{SBU}              \\
                            & BER$\downarrow$ & BER$\downarrow$ \\
  \midrule[1pt]
  SILT~\cite{ShadowDetection-SILT}                      & 0.011            & 0.044            \\
  SARA~\cite{ShadowDetection-SARA}                      & 0.018            & 0.029            \\
  \midrule[1pt]
  SAM (\PromptEverything)   & 0.205            & 0.256            \\
  SAM (\PromptBox)          & 0.150            & 0.141            \\
  SAM (\PromptPoint)        & 0.161            & 0.242            \\
  \midrule[1pt]
  SAM 2 (\PromptEverything) & 0.336            & 0.425            \\
  SAM 2 (\PromptBox)        & 0.180            & 0.153            \\
  SAM 2 (\PromptPoint)      & 0.220            & 0.273            \\
  \bottomrule[2pt]
\end{tabular}

%% file: table/image_transparent.tex
\begin{tabular}{l|c}
  \toprule[2pt]
                            & Trans10K~\cite{TransLab_Transparent}         \\
                            & BER$\downarrow$ \\
  \midrule[1pt]
  EBLNet~\cite{TOS-EBLNet}                    & 0.138            \\
  RFENet~\cite{TOS-RFENet}                    & 0.104            \\
  \midrule[1pt]
  SAM (\PromptEverything)   & 0.141            \\
  SAM (\PromptBox)          & 0.079            \\
  SAM (\PromptPoint)        & 0.057            \\
  \midrule[1pt]
  SAM 2 (\PromptEverything) & 0.231            \\
  SAM 2 (\PromptBox)        & 0.069            \\
  SAM 2 (\PromptPoint)      & 0.255            \\
  \bottomrule[2pt]
\end{tabular}

%% file: table/image_battery.tex
\begin{tabular}{lr|rr|rrr|rrr}
  \toprule[2pt]
   &                     & CFINet~\cite{CFINet} & MDCNet\cite{AI4Industry-PBD}~ & SAM (\PromptEverything) & SAM (\PromptBox) & SAM (\PromptPoint) & SAM 2 (\PromptEverything) & SAM 2 (\PromptBox) & SAM 2 (\PromptPoint) \\
  \midrule[1pt]
  \multirow{4}{*}{\rotatebox{90}{Regular}}
   & PN-ACC$\uparrow$   & 0.688  & 0.954  & \none                   & 0.147            & \none              & \none                     & 0.128              & \none                \\
   & AL-MAE$\downarrow$  & 4.022  & 2.337  & \none                   & 1.653            & 160.145            & \none                     & 1.318              & 212.986              \\
   & CL-MAE$\downarrow$  & 3.807  & 1.841  & \none                   & 1.983            & 516.093            & \none                     & 1.696              & 166.018              \\
   & OH-MAE$\downarrow$  & 3.950  & 2.042  & \none                   & 0.877            & \none              & \none                     & 1.311              & \none                \\
  \midrule[1pt]
  \multirow{4}{*}{\rotatebox{90}{Difficult}}
   & PN-ACC$\uparrow$   & 0.543  & 0.760  & \none                   & 0.133            & \none              & \none                     & 0.196              & \none                \\
   & AL-MAE$\downarrow$  & 4.960  & 2.440  & \none                   & 1.740            & 43.217             & \none                     & 1.601              & 101.321              \\
   & CL-MAE$\downarrow$  & 4.988  & 2.098  & \none                   & 2.338            & 301.011            & \none                     & 1.620              & 294.389              \\
   & OH-MAE$\downarrow$  & 3.977  & 2.109  & \none                   & 1.010            & \none              & \none                     & 1.330              & \none                \\
  \midrule[1pt]
  \multirow{4}{*}{\rotatebox{90}{Tough}}
   & PN-ACC$\uparrow$    & 0.328  & 0.512  & \none                   & \none            & 0.006              & \none                     & \none              & \none                \\
   & AL-MAE$\downarrow$  & 4.945  & 2.000  & \none                   & \none            & 551.057            & \none                     & \none              & 84.738               \\
   & CL-MAE$\downarrow$  & 4.662  & 1.465  & \none                   & 1.048            & 232.665            & \none                     & 1.174              & 63.642               \\
   & OH-MAE$\downarrow$  & 3.699  & 1.629  & \none                   & \none            & 48.528             & \none                     & \none              & \none                \\
  \bottomrule[2pt]
\end{tabular}

%% file: table/image_anomaly.tex
\begin{tabular}{lr|cc|ccc|ccc}
  \toprule[2pt]
   &                   & RD   & Patchcore & SAM  & SAM  & SAM  & SAM 2  & SAM 2 & SAM 2  \\
    &                   & ~\cite{AD-RD}   & ~\cite{AD-PatchCore} &  (\PromptEverything) &  (\PromptBox) &  (\PromptPoint) &  (\PromptEverything) & (\PromptBox) &  (\PromptPoint) \\
   
  \midrule[1pt]
  \multirow{5}{*}{\rotatebox{90}{MVTec~\cite{MVTec-AD}}}
   & I-AUROC$\uparrow$ & 98.6 & 99.2      & 55.1                    & 77.8             & 53.3               & 52.3                      & 72.7               & 94.9                 \\
   & I-AP$\uparrow$    & 99.5 & 99.8      & 75.0                    & 92.8             & 79.8               & 75.0                      & 91.6               & 97.7                 \\
   & P-AUROC$\uparrow$ & 97.8 & 99.4      & 51.1                    & 84.6             & 93.2               & 32.5                      & 84.5               & 97.8                 \\
   & P-AP$\uparrow$    & 58.0 & 56.1      & 4.9                     & 36.9             & 44.5               & 2.8                       & 28.8               & 78.4                 \\
   & P-PRO$\uparrow$   & 93.9 & 94.3      & 27.5                    & 62.8             & 75.7               & 13.7                      & 62.9               & 89.6                 \\
  \midrule[1pt]
  \multirow{5}{*}{\rotatebox{90}{VisA~\cite{VISAD-DS}}}
   & I-AUROC$\uparrow$ & 96.0 & 95.1      & 55.2                    & 95.8             & 45.7               & 54.4                      & 98.7               & 58.5                 \\
   & I-AP$\uparrow$    & 96.5 & 96.2      & 61.9                    & 98.2             & 56.0               & 61.8                      & 99.3               & 66.3                 \\
   & P-AUROC$\uparrow$ & 90.1 & 98.8      & 73.2                    & 87.6             & 53.3               & 43.1                      & 93.3               & 63.5                 \\
   & P-AP$\uparrow$    & 27.7 & 40.1      & 2.5                     & 54.2             & 1.0                & 1.1                       & 71.6               & 2.0                  \\
   & P-PRO$\uparrow$   & 70.9 & 91.2      & 35.8                    & 66.0             & 24.1               & 16.8                      & 83.4               & 25.7                 \\
  \midrule[1pt]
  \multirow{5}{*}{\rotatebox{90}{BTAD~\cite{BTAD}}}
   & I-AUROC$\uparrow$ & 93.7 & 94.7      & 75.5                    & 86.1             & 64.9               & 52.5                      & 81.0               & 71.8                 \\
   & I-AP$\uparrow$    & 98.5 & 98.9      & 66.5                    & 93.9             & 82.5               & 59.6                      & 90.4               & 82.2                 \\
   & P-AUROC$\uparrow$ & 95.8 & 97.8      & 47.8                    & 72.6             & 79.9               & 29.3                      & 76.5               & 79.1                 \\
   & P-AP$\uparrow$    & 51.7 & 52.0      & 3.5                     & 29.0             & 15.2               & 2.0                       & 30.7               & 47.9                 \\
   & P-PRO$\uparrow$   & 72.3 & 75.2      & 17.7                    & 41.8             & 49.7               & 4.2                       & 51.0               & 58.3                 \\
  \bottomrule[2pt]
\end{tabular}

%% file: table/image_medical.tex
\begin{tabular}{l|cc|cc|cc|cc}
  \toprule[2pt]
                                       &
  \multicolumn{2}{c|}{COVID-19~\cite{Inf-Net}} &
  \multicolumn{2}{c|}{BUSI~\cite{BUSI}}   &
  \multicolumn{2}{c|}{ISIC-2018~\cite{ISIC18}}     &
  \multicolumn{2}{c}{Polyp-Five}                                                                                                                                                        \\
   & Dice$\uparrow$ & mIoU$\uparrow$  & Dice$\uparrow$ & mIoU$\uparrow$  & Dice$\uparrow$ & mIoU$\uparrow$  & Dice$\uparrow$ & mIoU$\uparrow$  \\
  \midrule[1pt]
  InfNet~\cite{Inf-Net} &0.432&0.529&\none&\none&\none&\none&\none&\none\\
  DECORNet~\cite{LOS-DECOR-Net}&0.403&0.695&\none&\none&\none&\none&\none&\none\\
  AAUNet~\cite{LOS-AAU-Net}&\none&\none&0.475&0.652&\none&\none&\none&\none\\
  CMUNet~\cite{LOS-CMU-NeT}&\none&\none&0.545&0.830&\none&\none&\none&\none\\
  MALUNet~\cite{LOS-MALUNet}&\none&\none&\none&\none&0.863&0.854&\none&\none\\
  EGEUNet~\cite{LOS-EGE-UNet}&\none&\none&\none&\none&0.859&0.850&\none&\none\\
  LDNet~\cite{LOS-LDNet}&\none&\none&\none&\none&\none&\none&0.643&0.744\\
  WeakPolyp~\cite{LOS-WeakPolyp}&\none&\none&\none&\none&\none&\none&0.749&0.807\\
  \midrule[1pt]
  SAM (\PromptEverything)              & 0.431            & 0.705           & 0.477            & 0.670           & 0.350            & 0.548           & 0.486            & 0.705           \\
  SAM (\PromptBox)                     & 0.858            & 0.885           & 0.849            & 0.859           & 0.843            & 0.807           & 0.913            & 0.918           \\
  SAM (\PromptPoint)                   & 0.352            & 0.601           & 0.694            & 0.729           & 0.504            & 0.432           & 0.641            & 0.713           \\
  \midrule[1pt]
  SAM 2 (\PromptEverything)            & 0.244            & 0.612           & 0.156            & 0.528           & 0.182            & 0.467           & 0.109            & 0.520           \\
  SAM 2 (\PromptBox)                   & 0.893            & 0.909           & 0.895            & 0.896           & 0.824            & 0.816           & 0.927            & 0.928           \\
  SAM 2 (\PromptPoint)                 & 0.687            & 0.797           & 0.783            & 0.815           & 0.641            & 0.610           & 0.862            & 0.870           \\
  \bottomrule[2pt]
\end{tabular}

%% file: table/prompt20.tex
\begin{tabular}{l|cc|cc|c|c|cccccccc}
  \toprule[2pt]
                                    &
  \multicolumn{2}{c|}{Image SOD}    &
  \multicolumn{2}{c|}{Image COD}    &
  \multicolumn{1}{c|}{TOS}          &
  \multicolumn{1}{c|}{Image SD}     &
  \multicolumn{8}{c}{Image LOS}                                                                                                                                                                                                                                                                                                             \\
                                    &
  \multicolumn{2}{c|}{DUTS~\cite{DUTS}}         &
  \multicolumn{2}{c|}{COD10K~\cite{SINet_COD}}       &
  \multicolumn{1}{c|}{Trans10K~\cite{TransLab_Transparent}}     &
  \multicolumn{1}{c|}{SBU~\cite{SBU}}          &
  \multicolumn{2}{c}{COVID-19~\cite{Inf-Net}}      &
  \multicolumn{2}{c}{BUSI~\cite{BUSI}}          &
  \multicolumn{2}{c}{ISIC-2018~\cite{ISIC18}}     &
  \multicolumn{2}{c}{Polyp~\cite{PraNet}}                                                                                                                                                                                                                                                                                                                 \\
                                    & F$^{\omega}_{\beta}\uparrow$ & Sm$\uparrow$ & F$^{\omega}_{\beta}\uparrow$ & Sm$\uparrow$ & BER$\downarrow$ & BER$\downarrow$ & Dice$\uparrow$ & mIoU$\uparrow$ & Dice$\uparrow$ & mIoU$\uparrow$ & Dice$\uparrow$ & mIoU$\uparrow$ & Dice$\uparrow$ & mIoU$\uparrow$ \\
  \midrule[1pt]
  UniverSeg~\cite{UniverSeg}        & \none                        & \none        & \none                        & \none        & \none           & \none           & 0.673          & 0.368         & 0.775          & 0.600         & 0.761          & 0.708         & 0.553          & 0.261         \\
  SegGPT~\cite{SegGPT}              & 0.387                        & 0.628        & 0.404                        & 0.653        & 0.306           & 0.204           & 0.131          & 0.553         & 0.336          & 0.603         & 0.480          & 0.440         & 0.568          & 0.707         \\
  Spider~\cite{CDCU-Spider} & 0.882                        & 0.916        & 0.789                        & 0.867        & 0.055           & 0.027           & 0.696          & 0.813         & 0.838          & 0.866         & 0.894          & 0.874         & 0.824          & 0.866         \\
  SAM 2 (\PromptICL)                & 0.092                        & 0.478        & 0.429                        & 0.680        & 0.413           & 0.280           & 0.382          & 0.655         & 0.539          & 0.712         & 0.747          & 0.770         & 0.499          & 0.706         \\
  \bottomrule[2pt]
\end{tabular}

%% file: table/video_polyp.tex
\begin{tabular}{l|cc|cc|cc}
  \toprule[2pt]
                                            &
  \multicolumn{2}{c}{CVC-612-T~\cite{PNS-Net_Polyp}}    &
  \multicolumn{2}{|c}{CVC-612-V~\cite{PNS-Net_Polyp}}   &
  \multicolumn{2}{|c}{CVC-300-TV~\cite{PNS-Net_Polyp}}                                                          \\
                                            & Dice$\uparrow$ & mIoU$\uparrow$  & Dice$\uparrow$ & mIoU$\uparrow$  & Dice$\uparrow$ & mIoU$\uparrow$  \\
  \midrule[1pt]
  PNSNet~\cite{PNS-Net_Polyp}               & 0.841 & 0.788 & 0.859 & 0.804 & 0.863 & 0.805 \\
  M$^{2}$SNet~\cite{LOS-M2SNet}                      & 0.846 & 0.782 & 0.897 & 0.838 & 0.876 & 0.805 \\
  \midrule[1pt]
  SAM (\PromptEverything)                   & 0.622 & 0.768 & 0.432 & 0.681 & 0.412 & 0.677 \\
  SAM (\PromptBox)                          & 0.930 & 0.927 & 0.926 & 0.928 & 0.911 & 0.917 \\
  SAM (\PromptPoint)                        & 0.798 & 0.818 & 0.693 & 0.750 & 0.504 & 0.634 \\
  SAM (Propagated \PromptBox)               & 0.079 & 0.460 & 0.138 & 0.528 & 0.136 & 0.533 \\
  SAM (Propagated \PromptPoint)             & 0.518 & 0.584 & 0.321 & 0.472 & 0.166 & 0.421 \\
  \midrule[1pt]
  SAM 2 (1$\times$\PromptBox)                      & 0.798 & 0.866 & 0.762 & 0.846 & 0.897 & 0.906 \\
  SAM 2 (3$\times$\PromptBox)                      & 0.875 & 0.898 & 0.912 & 0.921 & 0.906 & 0.914 \\
  SAM 2 (5$\times$\PromptBox)                      & 0.909 & 0.925 & 0.920 & 0.928 & 0.914 & 0.920 \\
  \midrule[0.5pt]
  SAM 2 (1$\times$\PromptPoint)                    & 0.900 & 0.919 & 0.754 & 0.843 & 0.905 & 0.913 \\
  SAM 2 (3$\times$\PromptPoint)                    & 0.905 & 0.925 & 0.918 & 0.926 & 0.929 & 0.933 \\
  SAM 2 (5$\times$\PromptPoint)                    & 0.919 & 0.933 & 0.926 & 0.933 & 0.936 & 0.939 \\
  \midrule[0.5pt]
  SAM 2 (1$\times$\PromptMask)                     & 0.916 & 0.931 & 0.775 & 0.857 & 0.911 & 0.918 \\
  SAM 2 (3$\times$\PromptMask)                     & 0.915 & 0.933 & 0.930 & 0.937 & 0.944 & 0.947 \\
  SAM 2 (5$\times$\PromptMask)                     & 0.916 & 0.936 & 0.942 & 0.948 & 0.959 & 0.961 \\
  \bottomrule[2pt]
\end{tabular}

%% file: table/video_salient.tex
\begin{tabular}{l|cc|cc|cc|cc}
  \toprule[2pt]
                                    &
  \multicolumn{2}{c|}{DAVIS16~\cite{DAVIS16}}      &
  \multicolumn{2}{c|}{DAVSOD$_{E}$~\cite{DAVSOD}} &
  \multicolumn{2}{c|}{DAVSOD$_{N}$~\cite{DAVSOD}} &
  \multicolumn{2}{c}{DAVSOD$_{H}$~\cite{DAVSOD}}                                                                                                                                      \\
                                    & MAE$\downarrow$ & Sm$\uparrow$ & MAE$\downarrow$ & Sm$\uparrow$ & MAE$\downarrow$ & Sm$\uparrow$ & MAE$\downarrow$ & Sm$\uparrow$ \\
  \midrule[1pt]
  CoSTFormer~\cite{CoSTFormer}                        & 0.014           & 0.921        & 0.061           & 0.806        & 0.090           & 0.711        & \none           & \none        \\
  MAMNet~\cite{VideoSOD-MAMNet}                            & 0.020           & 0.897        & 0.065           & 0.777        & 0.088           & 0.688        & 0.089           & 0.622        \\
  \midrule[1pt]
  SAM (\PromptEverything)           & 0.013           & 0.899        & 0.038           & 0.808        & 0.055           & 0.750        & 0.027           & 0.791        \\
  SAM (\PromptBox)                  & 0.020           & 0.913        & 0.029           & 0.873        & 0.037           & 0.847        & 0.024           & 0.865        \\
  SAM (\PromptPoint)                & 0.009           & 0.927        & 0.030           & 0.879        & 0.044           & 0.828        & 0.030           & 0.828        \\
  SAM (Propagated \PromptBox)       & 0.042           & 0.712        & 0.091           & 0.614        & 0.108           & 0.587        & 0.098           & 0.544        \\
  SAM (Propagated \PromptPoint)     & 0.028           & 0.872        & 0.066           & 0.793        & 0.050           & 0.777        & 0.045           & 0.745        \\
  \midrule[1pt]
  SAM 2 (1$\times$\PromptBox)              & 0.006           & 0.936        & 0.051           & 0.773        & 0.055           & 0.769        & 0.062           & 0.652        \\
  SAM 2 (3$\times$\PromptBox)              & 0.008           & 0.933        & 0.044           & 0.813        & 0.054           & 0.787        & 0.049           & 0.731        \\
  SAM 2 (5$\times$\PromptBox)              & 0.007           & 0.936        & 0.044           & 0.812        & 0.047           & 0.805        & 0.050           & 0.687        \\
  \midrule[0.5pt]
  SAM 2 (1$\times$\PromptPoint)            & 0.005           & 0.949        & 0.043           & 0.800        & 0.062           & 0.736        & 0.060           & 0.645        \\
  SAM 2 (3$\times$\PromptPoint)            & 0.005           & 0.950        & 0.033           & 0.853        & 0.052           & 0.788        & 0.047           & 0.727        \\
  SAM 2 (5$\times$\PromptPoint)            & 0.005           & 0.954        & 0.027           & 0.872        & 0.041           & 0.820        & 0.039           & 0.754        \\
  \midrule[0.5pt]
  SAM 2 (1$\times$\PromptMask)             & 0.005           & 0.953        & 0.038           & 0.820        & 0.062           & 0.738        & 0.061           & 0.645        \\
  SAM 2 (3$\times$\PromptMask)             & 0.005           & 0.957        & 0.027           & 0.874        & 0.049           & 0.793        & 0.047           & 0.729        \\
  SAM 2 (5$\times$\PromptMask)             & 0.004           & 0.959        & 0.022           & 0.882        & 0.040           & 0.824        & 0.039           & 0.761        \\
  \bottomrule[2pt]
\end{tabular}

%% file: table/video_camo.tex
\begin{tabular}{l|cc|cc}
  \toprule[2pt]
                                                       &
  \multicolumn{2}{c|}{MoCA-Mask~\cite{VCOD-MoCA-Mask}} &
  \multicolumn{2}{c}{CAD~\cite{VCOD-CAD}}                                                                               \\
                                                       & MAE$\downarrow$ & Sm$\uparrow$ & MAE$\downarrow$ & Sm$\uparrow$ \\
  \midrule[1pt]
  SLT-Net~\cite{SLT-Net}                               & 0.027           & 0.637        & 0.031           & 0.696        \\
  ZoomNext~\cite{Zoomnext}                             & 0.010           & 0.734        & 0.020           & 0.757        \\
  \midrule[1pt]

  SAM (\PromptEverything)                              & 0.010           & 0.638        & 0.019           & 0.735        \\
  SAM (\PromptBox)                                     & 0.005           & 0.817        & 0.017           & 0.851        \\
  SAM (\PromptPoint)                                   & 0.025           & 0.791        & 0.033           & 0.793        \\
  SAM (Propagated \PromptBox)                          & 0.011           & 0.660        & 0.053           & 0.560        \\
  SAM (Propagated \PromptPoint)                        & 0.074           & 0.604        & 0.103           & 0.551        \\
  \midrule[1pt]
  SAM 2 (1$\times$\PromptBox)                          & 0.006           & 0.790        & 0.012           & 0.862        \\
  SAM 2 (3$\times$\PromptBox)                          & 0.005           & 0.798        & 0.009           & 0.874        \\
  SAM 2 (5$\times$\PromptBox)                          & 0.005           & 0.810        & 0.009           & 0.874        \\
  \midrule[1pt]
  SAM 2 (1$\times$\PromptPoint)                        & 0.004           & 0.803        & 0.009           & 0.857        \\
  SAM 2 (3$\times$\PromptPoint)                        & 0.003           & 0.829        & 0.008           & 0.863        \\
  SAM 2 (5$\times$\PromptPoint)                        & 0.003           & 0.840        & 0.007           & 0.875        \\
  \midrule[1pt]
  SAM 2 (1$\times$\PromptMask)                         & 0.004           & 0.820        & 0.008           & 0.883        \\
  SAM 2 (3$\times$\PromptMask)                         & 0.003           & 0.844        & 0.006           & 0.900        \\
  SAM 2 (5$\times$\PromptMask)                         & 0.002           & 0.860        & 0.005           & 0.913        \\
  \bottomrule[2pt]
\end{tabular}

%% file: table/video_shadow.tex
\begin{tabular}{l|c|c}
  \toprule[2pt]
                                &VISAD-DS~\cite{VISAD-DS}       & VISAD-MOS~\cite{VISAD-DS}       \\
                                & BER$\downarrow$ & BER$\downarrow$ \\
  \midrule[1pt]
  SANet~\cite{VideoShadowDetection-SANet}                        & 0.131           & 0.259           \\
  \midrule[1pt]
  SAM (\PromptEverything)       & 0.146           & 0.342           \\
  SAM (\PromptBox)              & 0.091           & 0.125           \\
  SAM (\PromptPoint)            & 0.135           & 0.266           \\
  SAM (Propagated \PromptBox)   & 0.183           & 0.346           \\
  SAM (Propagated \PromptPoint) & 0.287           & 0.342           \\
  \midrule[1pt]
  SAM 2 (1$\times$\PromptBox)          & 0.292           & 0.406           \\
  SAM 2 (3$\times$\PromptBox)          & 0.283           & 0.367           \\
  SAM 2 (5$\times$\PromptBox)          & 0.250           & 0.349           \\
  \midrule[0.5pt]
  SAM 2 (1$\times$\PromptPoint)        & 0.136           & 0.351           \\
  SAM 2 (3$\times$\PromptPoint)        & 0.113           & 0.307           \\
  SAM 2 (5$\times$\PromptPoint)        & 0.104           & 0.335           \\
  \midrule[0.5pt]
  SAM 2 (1$\times$\PromptMask)         & 0.106           & 0.317           \\
  SAM 2 (3$\times$\PromptMask)         & 0.091           & 0.210           \\
  SAM 2 (5$\times$\PromptMask)         & 0.070           & 0.172           \\
  \bottomrule[2pt]
\end{tabular}

%% file: table/3d_medical_brats2020.tex
\begin{tabular}{l|ccc|ccc|ccc|ccc}
    \toprule[2pt]
                               &
    \multicolumn{3}{c|}{Flair} &
    \multicolumn{3}{c|}{T1ce}  &
    \multicolumn{3}{c|}{T1}    &
    \multicolumn{3}{c}{T2}                                                                                                     \\
                               &
    Dice$_{WT}\uparrow$        &
    Dice$_{TC}\uparrow$        &
    Dice$_{ET}\uparrow$        &
    Dice$_{WT}\uparrow$        &
    Dice$_{TC}\uparrow$        &
    Dice$_{ET}\uparrow$        &
    Dice$_{WT}\uparrow$        &
    Dice$_{TC}\uparrow$        &
    Dice$_{ET}\uparrow$        &
    Dice$_{WT}\uparrow$        &
    Dice$_{TC}\uparrow$        &
    Dice$_{ET}\uparrow$                                                                                                        \\
    \midrule[1pt]
    3D U-Net~\cite{3DUNet}     & 0.900 & 0.807 & 0.792 & 0.900 & 0.807 & 0.792 & 0.900 & 0.807 & 0.792 & 0.900 & 0.807 & 0.792 \\
    EoFormer~\cite{EoFormer}   & 0.908 & 0.864 & 0.832 & 0.908 & 0.864 & 0.832 & 0.908 & 0.864 & 0.832 & 0.908 & 0.864 & 0.832 \\
    \midrule[1pt]
    SAM 2 (1$\times$\PromptBox)       & 0.566 & 0.574 & 0.579 & 0.566 & 0.574 & 0.582 & 0.566 & 0.570 & 0.579 & 0.566 & 0.574 & 0.579 \\
    SAM 2 (3$\times$\PromptBox)       & 0.560 & 0.574 & 0.579 & 0.566 & 0.574 & 0.582 & 0.566 & 0.570 & 0.618 & 0.560 & 0.574 & 0.579 \\
    SAM 2 (5$\times$\PromptBox)       & 0.555 & 0.574 & 0.670 & 0.567 & 0.615 & 0.582 & 0.566 & 0.578 & 0.579 & 0.555 & 0.574 & 0.670 \\
    \midrule[0.5pt]
    SAM 2 (1$\times$\PromptPoint)     & 0.700 & 0.683 & 0.643 & 0.681 & 0.684 & 0.592 & 0.675 & 0.615 & 0.558 & 0.700 & 0.683 & 0.550 \\
    SAM 2 (3$\times$\PromptPoint)     & 0.788 & 0.753 & 0.700 & 0.747 & 0.780 & 0.707 & 0.761 & 0.728 & 0.659 & 0.788 & 0.753 & 0.650 \\
    SAM 2 (5$\times$\PromptPoint)     & 0.835 & 0.793 & 0.723 & 0.804 & 0.794 & 0.733 & 0.799 & 0.748 & 0.676 & 0.835 & 0.793 & 0.670 \\
    \midrule[0.5pt]
    SAM 2 (1$\times$\PromptMask)      & 0.706 & 0.698 & 0.645 & 0.700 & 0.676 & 0.639 & 0.670 & 0.639 & 0.587 & 0.706 & 0.698 & 0.647 \\
    SAM 2 (3$\times$\PromptMask)      & 0.864 & 0.803 & 0.760 & 0.841 & 0.808 & 0.775 & 0.810 & 0.765 & 0.734 & 0.864 & 0.803 & 0.760 \\
    SAM 2 (5$\times$\PromptMask)      & 0.889 & 0.843 & 0.796 & 0.871 & 0.838 & 0.803 & 0.860 & 0.815 & 0.783 & 0.889 & 0.843 & 0.796 \\
    \bottomrule[2pt]
\end{tabular}

%% file: table/3d_medical_isbi2015.tex
\begin{tabular}{l|c}
    \toprule[2pt]
                            & Flair               \\
                            & Dice$_{MS}\uparrow$ \\
    \midrule[1pt]
    DRU-Net~\cite{ISBI2015-DRU-Net}   & 0.663               \\
    AttU-Net~\cite{ISBI2015-AttnUNet} & 0.803               \\
    \midrule[1pt]
    SAM 2 (1$\times$\PromptBox)    & 0.635               \\
    SAM 2 (3$\times$\PromptBox)    & 0.638               \\
    SAM 2 (5$\times$\PromptBox)    & 0.636               \\
    \midrule[0.5pt]
    SAM 2 (1$\times$\PromptPoint)  & 0.630               \\
    SAM 2 (3$\times$\PromptPoint)  & 0.630               \\
    SAM 2 (5$\times$\PromptPoint)  & 0.630               \\
    \midrule[0.5pt]
    SAM 2 (1$\times$\PromptMask)   & 0.728               \\
    SAM 2 (3$\times$\PromptMask)   & 0.763               \\
    SAM 2 (5$\times$\PromptMask)   & 0.768               \\
    \bottomrule[2pt]
\end{tabular}

%% file: table/image_robustness.tex
\begin{tabular}{l|cc|ccccc}
  \toprule[2pt]
                                        &
  \multicolumn{2}{c|}{DUTS~\cite{DUTS}} &
  \multicolumn{5}{c}{MVTec-AD~\cite{MVTec-AD}}                                                                                                                                                                                                                                  \\
                                        & F$^{\omega}_{\beta}\uparrow$   & Sm$\uparrow$                   & I-AUROC$\uparrow$              & I-AP$\uparrow$                 & P-AUROC$\uparrow$              & P-AP$\uparrow$                  & P-PRO$\uparrow$                \\
  \midrule[1pt]
  SAM (\PromptBox)                      & 0.894$_{\pm \text{1.1E-03}}$   & 0.888$_{\pm \text{1.1E-03}}$   & 0.726$_{\pm \text{7.0E-04}}$   & 0.917$_{\pm \text{2.00E-04}}$  & 0.845$_{\pm \text{1.0E-04}}$   & 0.289$_{\pm \text{2.0E-04}}$    & 0.630$_{\pm \text{2.0E-04}}$   \\
  \quad$\Delta$                         & \basevalue{$\downarrow$2.75\%} & \basevalue{$\downarrow$2.35\%} & \basevalue{$\downarrow$6.66\%} & \basevalue{$\downarrow$1.22\%} & \basevalue{$\downarrow$0.11\%} & \basevalue{$\downarrow$21.73\%} & \basevalue{$\uparrow$0.29\%}   \\
  SAM (\PromptPoint)                    & 0.831$_{\pm \text{2.3E-03}}$   & 0.852$_{\pm \text{2.3E-03}}$   & 0.585$_{\pm \text{6.3E-03}}$   & 0.821$_{\pm \text{2.4E-03}}$   & 0.924$_{\pm \text{1.9E-03}}$   & 0.430$_{\pm \text{8.7E-03}}$    & 0.731$_{\pm \text{1.8E-03}}$   \\
  \quad$\Delta$                         & \basevalue{$\downarrow$6.30\%} & \basevalue{$\downarrow$4.01\%} & \basevalue{$\uparrow$9.76\%}   & \basevalue{$\uparrow$2.87\%}   & \basevalue{$\downarrow$0.88\%} & \basevalue{$\downarrow$3.46\%}  & \basevalue{$\downarrow$3.50\%} \\
  \midrule[1pt]
  SAM 2 (\PromptBox)                    & 0.857$_{\pm \text{2.3E-03}}$   & 0.864$_{\pm \text{1.3E-03}}$   & 0.779$_{\pm \text{5.0E-04}}$   & 0.929$_{\pm \text{2.00E-04}}$  & 0.847$_{\pm \text{1.0E-04}}$   & 0.369$_{\pm \text{1.0E-04}}$    & 0.629$_{\pm \text{1.0E-04}}$   \\
  \quad$\Delta$                         & \basevalue{$\downarrow$7.75\%} & \basevalue{$\downarrow$6.09\%} & \basevalue{$\uparrow$7.11\%}   & \basevalue{$\uparrow$1.38\%}   & \basevalue{$\uparrow$0.22\%}   & \basevalue{$\uparrow$28.09\%}   & \basevalue{$\downarrow$0.08\%} \\
  SAM 2 (\PromptPoint)                  & 0.773$_{\pm \text{1.4E-03}}$   & 0.792$_{\pm \text{9.0E-04}}$   & 0.939$_{\pm \text{6.7E-03}}$   & 0.970$_{\pm \text{3.4E-03}}$   & 0.975$_{\pm \text{1.0E-03}}$   & 0.774$_{\pm \text{3.9E-03}}$    & 0.878$_{\pm \text{2.2E-03}}$   \\
  \quad$\Delta$                         & \basevalue{$\downarrow$4.20\%} & \basevalue{$\downarrow$2.77\%} & \basevalue{$\downarrow$1.05\%} & \basevalue{$\downarrow$0.70\%} & \basevalue{$\downarrow$0.27\%} & \basevalue{$\downarrow$1.34\%}  & \basevalue{$\downarrow$1.96\%} \\
  \bottomrule[2pt]
\end{tabular}

%% file: table/video_robustness.tex
\begin{tabular}{l|cc}
  \toprule[2pt]
  &
  \multicolumn{2}{c}{DAVIS16~\cite{DAVIS16}}                                                                                                                                                                                 \\
                                          & MAE$\downarrow$                 & Sm$\uparrow$                   \\
  \midrule[1pt]
  SAM (\PromptBox)                       
  & 0.021$_{\pm \text{1.2E-03}}$    
  & 0.905$_{\pm \text{2.1E-03}}$   \\
  \quad$\Delta$       
  & \basevalue{$\downarrow$2.96\%}  
  & \basevalue{$\downarrow$0.88\%} \\
  SAM (\PromptPoint)          
  & 0.013$_{\pm \text{1.2E-03}}$ 
  & 0.916$_{\pm \text{2.1E-03}}$   \\
  \quad$\Delta$                     
  & \basevalue{$\downarrow$41.76\%} 
  & \basevalue{$\downarrow$1.24\%} \\
  \midrule[1pt]
  SAM 2 (\PromptBox)            
  & 0.007$_{\pm \text{1.2E-03}}$    
  & 0.930$_{\pm \text{6.5E-03}}$   \\
  \quad$\Delta$                             
  & \basevalue{$\downarrow$17.46\%} 
  & \basevalue{$\downarrow$0.60\%} \\
  SAM 2 (\PromptPoint)                
  & 0.006$_{\pm \text{1.7E-03}}$   
  & 0.946$_{\pm \text{5.2E-03}}$   \\
  \quad$\Delta$                   
  & \basevalue{$\downarrow$15.38\%} 
  & \basevalue{$\downarrow$0.32\%} \\
  SAM 2 (\PromptMask)                   
  & 0.008$_{\pm \text{1.0E-03}}$    
  & 0.934$_{\pm \text{5.3E-03}}$   \\
  \quad$\Delta$                    
  & \basevalue{$\downarrow$59.62\%} 
  & \basevalue{$\downarrow$2.05\%} \\
  \bottomrule[2pt]
\end{tabular}